\documentclass{article}
\usepackage[eatrack]{log_2023}				

\usepackage{booktabs}						
\usepackage{multirow}						
\usepackage{amsfonts}						
\usepackage{graphicx}						
\usepackage{duckuments}						
\usepackage{caption}
\usepackage{subcaption}
\usepackage{bm}
\usepackage[toc,page]{appendix}

\usepackage[numbers,compress,sort]{natbib}	

\title[Triplet Edge Attention for Algorithmic Reasoning]{Triplet Edge Attention for Algorithmic Reasoning}

\author[Yeonjoon Jung and Sungsoo Ahn]{%
Yeonjoon Jung\\
{Pohang University of Science and Technology}\\
\email{jeffjung01@postech.ac.kr}\And
Sungsoo Ahn\\
{Pohang University of Science and Technology}\\
\email{sungsoo.ahn@postech.ac.kr}
}

\begin{document}

\maketitle

\begin{abstract}
This work investigates neural algorithmic reasoning to develop neural networks capable of learning from classical algorithms. The main challenge is to develop graph neural networks that are expressive enough to predict the given algorithm outputs while generalizing well to out-of-distribution data. In this work, we introduce a new graph neural network layer called Triplet Edge Attention (TEA), an edge-aware graph attention layer. Our algorithm works by precisely computing edge latent, aggregating multiple triplet messages using edge-based attention. We empirically validate our TEA layer in the CLRS benchmark and demonstrate a $5\%$ improvement on average. In particular, we achieve a $30\%$ improvement for the string algorithms compared to the state-of-the-art model.
\end{abstract}

\section{Introduction}
Neural networks are undergoing rapid development and unprecedented performance in tackling intricate tasks across a wide range of domains. However, a critical vulnerability of neural networks lies in their robustness, which is the capacity to maintain consistent performance on all inputs \cite{carlini:2022towards}. Despite their high accuracy on the training distribution, neural networks often exhibit subpar performance during the inference stage, particularly when confronting input data lying outside the training data distribution. Neural algorithmic reasoning \cite{velivckovic:2021neural} is a promising avenue to address this robustness problem by combining neural network models with classical algorithms, which entails training the algorithm itself to the neural network. 

In contrast to traditional algorithmic solutions for real-world tasks, which embed the task into a known domain and then select an algorithm to solve the proxy-domain problem, neural algorithmic reasoning allows direct input without the proxy step \cite{velivckovic:2021neural}. Furthermore, neural algorithmic networks can be repurposed as a pretrained model without losing their generality. This versatility enables the execution of complex tasks by synthesizing multiple algorithmic reasoning layers and even discovering new algorithms from unconventional perspectives compared to human analyses~\cite{velivckovic:2022clrs}. 

Algorithmic reasoning tasks are commonly accomplished using graph neural networks (GNNs)~\cite{scarselli:2008gnn}, which excel in terms of expression power compared to models using sequential or grid-based inputs. Corresponding attribute facilitates the efficient expression of algorithms \cite{velivckovic:2019neural} such as Breadth-First Search \cite{moore:1959shortest} or Bellman-Ford \cite{bellman:1958routing}. This research aims to develop a graph-based algorithmic learner, a GNN capable of learning various algorithms involving diverse input types. The previous state-of-the-art (SOTA) model, i.e., Triplet-GMPNN \citep{ibarz:2022genaralist}, displayed remarkable enhancements across the CLRS benchmark~\cite{velivckovic:2022clrs} thanks to its capability to reason over triplet of vertices. 

\textbf{Contribution.} In this work, we focus on refining neural architectures to better align with the algorithmic reasoning tasks. To this end, we devise a novel edge-attention method tailored for reasoning, coined \textbf{T}riplet \textbf{E}dge \textbf{A}ttention (\textbf{TEA}), an efficient method for computing edge latent suitable for both node-based and edge-based algorithms. We use the newly proposed attention to construct \textbf{T}riplet \textbf{E}dge \textbf{A}ttention \textbf{M}essage passing neural network (\textbf{TEAM}), which we combine with the encoder-processor-decoder network to enhance the computation of algorithmic outputs. We evaluate our algorithm on the CLRS-30 benchmark \citep{velivckovic:2022clrs} and achieve state-of-the-art results with an average rank of $1.63$. 




\section{Preliminaries}
\paragraph{Problem setting} 
Algorithms solve a particular problem in a finite amount of time with unambiguous and executable steps. Several algorithms in the CLRS-30 benchmark, such as the Matrix-Chain-Order algorithm \citep{aho:1974design} or the Floyd-Warshall algorithm \citep{floyd:1962algorithm}, require edge-based reasoning to determine the next states of algorithms. Thus, given graph $G=(\mathcal{V},\mathcal{E})$ with vertices $\mathcal{V}$ and edges $\mathcal{E}$, handling a $|\mathcal{V}|^{3}$ message is beneficial for edge-level reasoning by aggregating intermediate node effects \cite{dudzik:2022graph}. 

\paragraph{Hints}
Hints are time series data that represent algorithm states, which contain the necessary information needed to replicate the logical, step-by-step execution of algorithms \cite{velivckovic:2022clrs}. The type of hints depends on the task and can be stored at node, edge, or graph level. In the absence of monitoring the model with ground-truth hints, it is difficult to confirm that the intended algorithm is being learned during the training. For example, when an insertion sort algorithm is given as input, the model might learn other behaviors such as heap sort \cite{williams:1964heapsort} or quicksort \cite{hoare:1962quicksort} algorithm without the help of hints. To address this issue, we adopted an architecture that anticipates the algorithm's hint trajectory through an iterative process before generating the final output.

\paragraph{Encode-process-decode paradigm}
We embrace the encode-process-decode paradigm as presented in the CLRS benchmark \citep{velivckovic:2022clrs}. For a given task $\tau$, at each $t$-th time step, the encoder $f_{\tau}$ performs linear encoding for input and hint, embedding them as high-dimensional vectors. These embeddings of inputs and hints located in the nodes all have the same dimension and are added together; the same happens with hints and inputs located in the edges, and in the graph. Thus, at the end of the encoding step for a time step $t$ of the algorithm, we have a single set of embeddings $\{\mathbf{x}_{i}^{(t)}, \mathbf{e}_{ij}^{(t)}, \mathbf{g}^{(t)}\}$ with $\mathbf{x}_{i}^{(t)} \in \mathbb{R}^{n\times h}$, $\mathbf{e}_{ij}^{(t)} \in \mathbb{R}^{n\times n}$, 
$\mathbf{g}^{(t)} \in \mathbb{R}^{h}$, in the nodes, edges, and graphs, respectively. Note that this process remains independent of the size or type of the inputs and hints inherent to a particular algorithm, allowing us to share this latent space across all the thirty algorithms in CLRS. 

 
 

\section{Triplet Edge Attention}
We propose a novel \textbf{T}riplet \textbf{E}dge \textbf{A}ttention (\textbf{TEA}) which captures multiple node features to compute the edge latent values. Previous attempts for graph attention were made in both node-based and edge-based level \citep{velickovic:2017graph, wang:2019improving, wang:2020multi, brody:2021attentive, chen:2021edgefeatured, lee:2023towards}. Our TEA computes the edge latent representation $\mathbf{h}_{ij}$ as follows: 
\begin{equation}
    \mathbf{h}_{ij} = \operatorname{ReLU}\left(\sum_{k\in{\mathcal{N}_{i}}\cup{\mathcal{N}_{j}}}\bm{\alpha}_{ijk}\left({\mathbf{W}' \mathbf{e}_{ik}}\right)\right), \label{eq:xx7}
\end{equation}
where $\bm{\alpha}_{ijk}$ is the newly proposed edge-based triplet attention computed as follows:
\begin{align} 
    \mathbf{t}'_{ijk} &= \mathbf{a}^T\operatorname{LeakyReLU}(\mathbf{t}_{ijk}), \qquad 
    \bm{\alpha}_{ijk} = \frac{\exp(\mathbf{t}'_{ijk})}{\sum_{k'\in{\mathcal{N}_{i}}\cup{\mathcal{N}_{j}}}\exp(\mathbf{t}'_{ijk'})}, \\ 
    \mathbf{t}_{ijk} & = \mathbf{W} [\mathbf{x}_{i} \parallel \mathbf{x}_{j} \parallel \mathbf{x}_{k} \parallel \mathbf{e}_{ij} \parallel \mathbf{e}_{ik} \parallel \mathbf{e}_{jk} \parallel \mathbf{g}]. \label{eq:xx4}
\end{align}
Note that the multi-head attention can also be implemented as follows: 
\begin{equation}
    \mathbf{h}_{ij} = \underset{m=1}{\overset{M}{\Big\Vert}} \left( \operatorname{ReLU} \left( \sum_{k\in{\mathcal{N}_{i}}\cup{\mathcal{N}_{j}}}\bm{\alpha}_{ijk}^m\left({\mathbf{W}^m  \mathbf{e}_{ik}} \right) \right) \right) \label{eq:xx8}
\end{equation}

where, $\parallel$ indicates concatenation, and the activation function from Equation \eqref{eq:xx7}, \eqref{eq:xx8} may be changed depending on the task. The major difference of previous triplet reasoning and our TEA is that, we efficiently map representations for all additional nodes through edge-level attention, instead of maximization over triplet messages.
The computational complexity of a single attention head TEA layer is expressed as $\mathbf{O}\left( |\mathcal{V}|^{2}h^{2} + |\mathcal{V}|^{3}h\right)$ which matches the time complexity of triplet reasoning method, where $h$ is the hint dimension. This enables fair comparison between TEA and triplet reasoning based models respective to its computational cost. Practical ratios of the number of parameters are stated in \ref{tab:params}.

Our TEA method exhibits effectiveness not only for edge-based reasoning problems but also for node-based reasoning problems. Compared to the traditional message passing with two steps, where the effects of 2-hop neighbors are indirectly computed through 1-hop nodes, TEA directly calculates the influence of 2-hop neighbors relative to the target node's state. This direct calculation ensures accurate attention score computation. The TEA method is also beneficial when dealing with fully-connected graphs, which aligns with the practical condition of our experiment. Contrary to the two-step message passing or two-step graph attention, which stores information as node latents, the TEA method preserves computed information as edge latents. This distinction becomes critical within a fully-connected graph, where all node latents are used to predict the output of each node. In contrast, our edge latent solely affects the two connected nodes during the final step. This difference in the influence scope allows precise reasoning for individual nodes and the aggregation of multiple conditions to determine the next execution state of the algorithm.

\begin{figure}[t]
    \begin{subfigure}[b]{0.45\linewidth}
        \centering
        \includegraphics[height=0.4\linewidth]{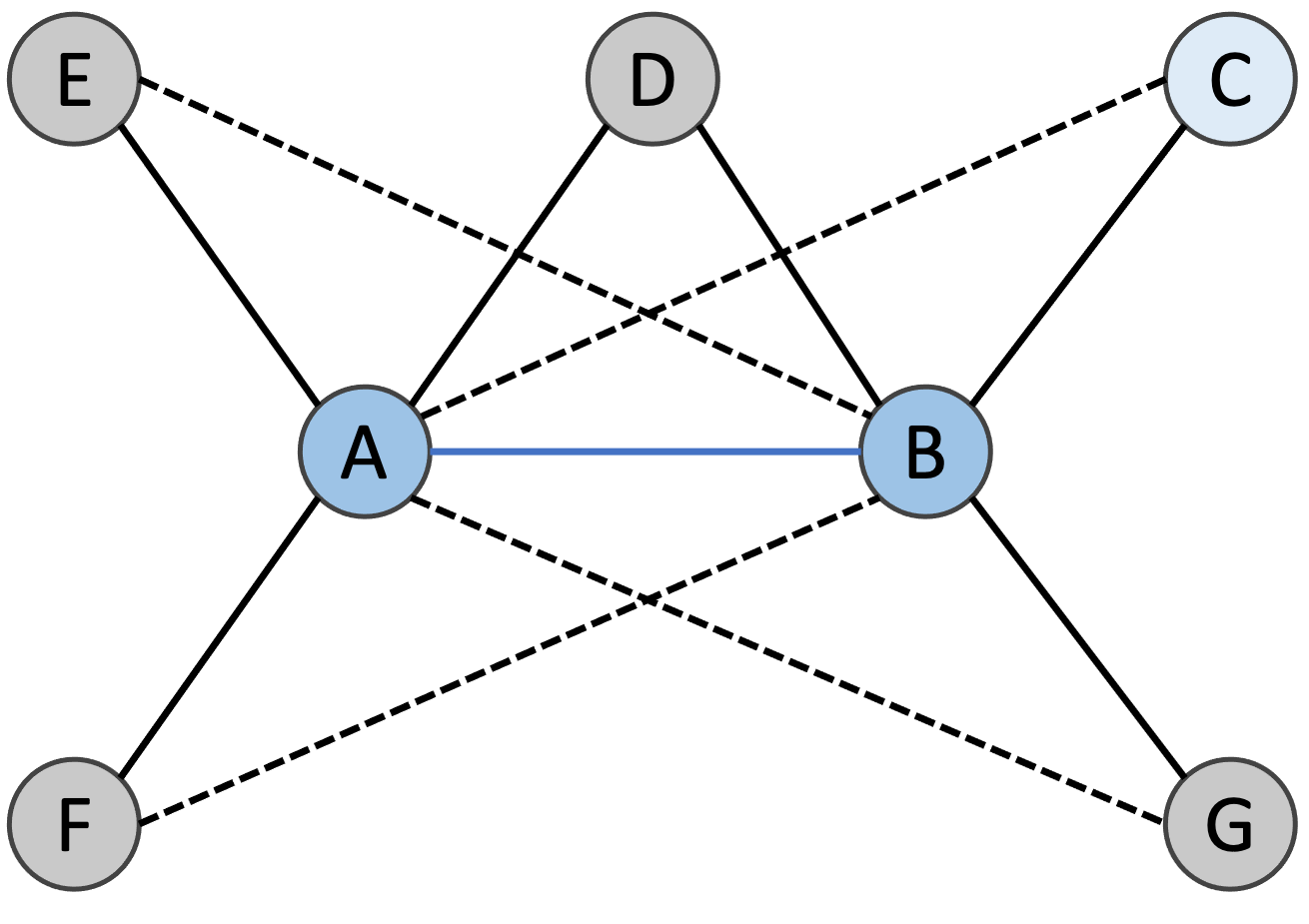}
	\caption{TEA for edge ${AB}$}
	\label{fig:TEA_1}
    \end{subfigure}
    \begin{subfigure}[b]{0.45\linewidth}
        \centering
        \includegraphics[height=0.4\linewidth]{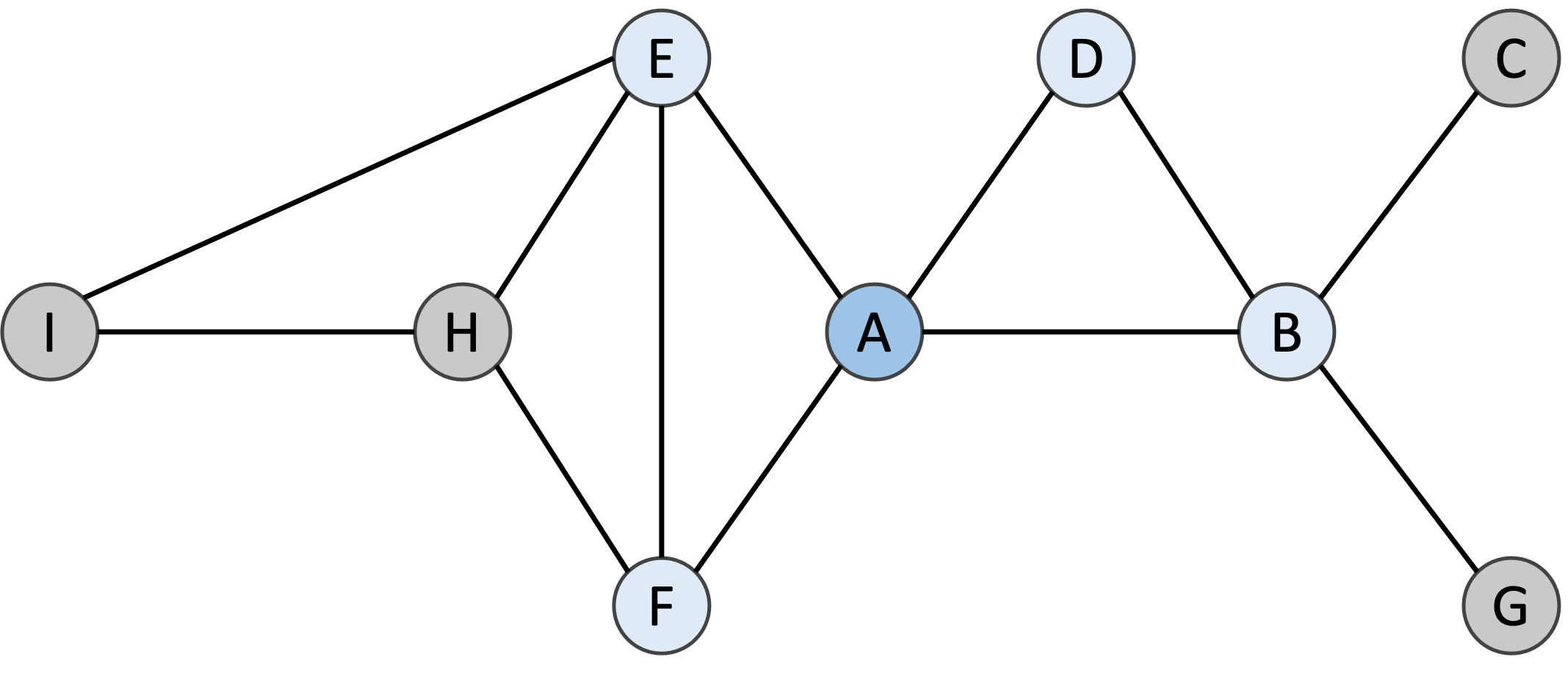}
	\caption{Final Computation for node $A$}
	\label{fig:TEA_2}
    \end{subfigure}
        \caption{Triplet Edge Attention structure for edge latent $AB$ and final computation process for Node $A$'s Output. A solid line represents an edge, and a dotted line implies that the nodes are not directly connected but are still considered for TEA. Colored nodes from \ref{fig:TEA_2} are $A$'s neighbors which are considered for the output computation.}
        \label{fig:TEA}
\end{figure}

We demonstrate the effectiveness of our TEA in the example of Figure \ref{fig:TEA_1}. Here, the effect of node ${C}$ is embedded in the edge latent $\mathbf{h}_{AB}$ considering features of nodes $A$, ${B}$, ${C}$, edges ${AB}$, ${BC}$, ${AC}$, and graph $G$. It is noteworthy that edge ${AC}$ could be a zero vector if the two nodes are not connected. In the case we are computing the output for node $A$, as shown in Figure \ref{fig:TEA_2}, the traditional message passing method with two steps would embed the effect of node ${C}$ into node ${B}$, then propagate it to node $A$. On the other hand, even when nodes $A$ and ${C}$ are not directly linked, our TEA method computes the influence of node ${C}$ relative to the state of node $A$, enabling accurate computation of attention score.

\section{Experiment}
\subsection{Implementation}
Our experiments were conducted using the CLRS benchmark \cite{velivckovic:2022clrs}, which includes 30 classical algorithms categorized into 8 distinct groups, reformatted as graph structure enabling training on GNNs. During the training process, we applied model improvement techniques \cite{ibarz:2022genaralist} to enhance the model's robustness. Additionally, we replaced the in-distribution validation set of size 16 into a OOD data of size 32, to prevent over-fitting on a particular graph size and assure the model is learning the general algorithm. We also applied OOD validation to the previous SOTA model to make comparison of its effect. The effect of changing validation set is examined in the results section. For the test set, OOD data of size 64 were used. 

We use the Triplet-GMPNN architecture \cite{ibarz:2022genaralist} as a baseline model for this research. The model is designed with encoder-processor-decoder architecture for hint-based tasks introduced from the previous work \cite{velivckovic:2022clrs}. The processor network for Triplet-GMPNN is a fully connected MPNN \cite{gilmerz:2017mpnn} with triplet reasoning and gating method \cite{ibarz:2022genaralist}. 


\subsection{Results}

\begin{figure}[t]
	\centering
	\includegraphics[width=1\linewidth, height=0.25\linewidth]{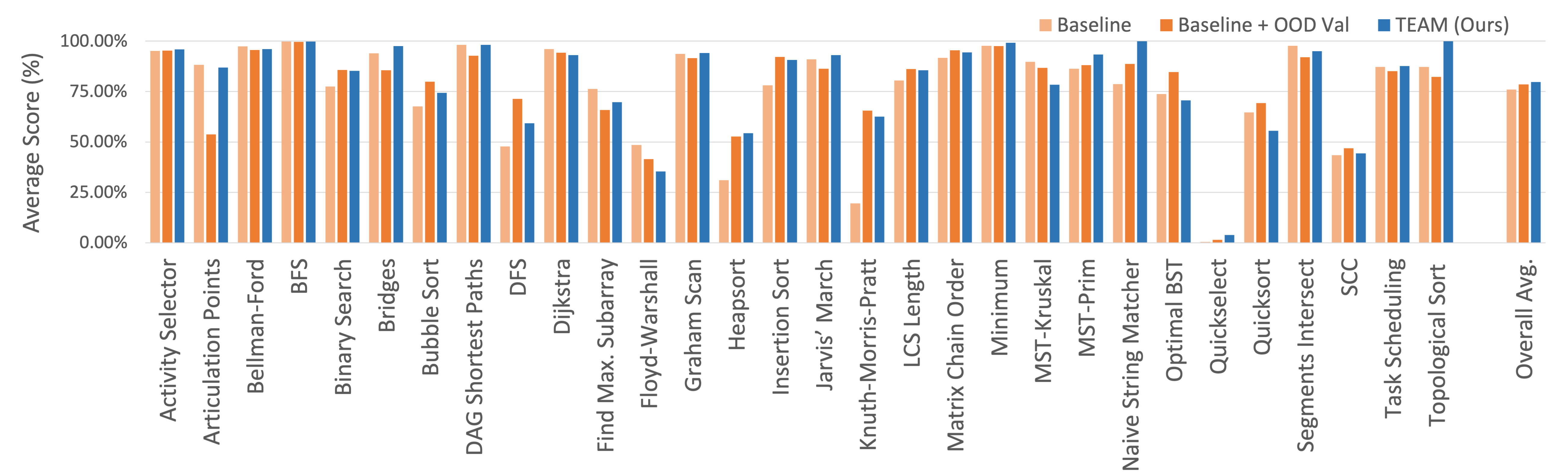}
	\caption{Bar chart for OOD micro-F1 scores of baseline (Triplet-GMPNN), baseline with OOD validation set, and our best model TEAM, after 10,000 training steps on each algorithm.}
	\label{fig:micro_f1}
\end{figure}

We evaluated OOD performance over CLRS-30 benchmark \cite{velivckovic:2022clrs}, with 5 repetitions. Table \ref{tab:category}, \ref{tab:30} and Figure \ref{fig:micro_f1} present the micro-F1 score for OOD test data, comparing the baseline Triplet-GMPNN \cite{ibarz:2022genaralist}, the baseline with OOD validation set, and our TEAM model. TEAM model exhibited the highest performance, showing a 5.09\% improvement compared to the previous SOTA, while the appliance of OOD validation in the baseline model yielded a 3.51\% improvement. Additionally, our TEAM model outperformed the baseline across 6 algorithm categories, displaying the highest rank among models, suggesting that TEAM shows the most robust performance overall in all algorithm categories. 

Significant performance increases were observed in sorting algorithms (insertion sort, bubble sort, heapsort~\cite{williams:1964heapsort}, quicksort~\cite{hoare:1962quicksort}) and string algorithms (Naive string matching, Knuth-Morris-Pratt (KMP) string matcher~\cite{knuth:1977strings}). In particular, our TEAM model achieved an 81.24\% performance for string algorithms requiring multiple comparison capability, a considerable improvement over the previous best of 49.09\%.

The appliance of the new training framework with the OOD validation set improved the baseline model in terms of overall average score and average rank. It also increased the number of algorithms solvable exceeding 50\% performance compared to the baseline. Whereas, our TEAM model, also using the OOD validation set, showed the best result for all standards including average performance, average rank, and the number of algorithms over 50\% \& 90\% OOD performance. 
A notable observation is the large intersection of algorithms surpassing the baseline between the baseline with OOD validation and our TEAM model. This implies that the effectiveness of the OOD validation method might be dependent on the target task's characteristics.

\begin{table}[t]
	\centering
	\caption{Average OOD micro-F1 scores of Memnet, MPNN, PGN, baseline (Triplet-GMPNN), baseline with OOD validation set, and our best TEAM model, after 10,000 training steps on each algorithm category.}
	\label{tab:category}
    \resizebox{\textwidth}{!}{
	\begin{tabular}{lcccccc}
		\toprule
		Algorithm & Memnet~\cite{velivckovic:2022clrs} & MPNN~\cite{velivckovic:2022clrs} & PGN~\cite{velivckovic:2022clrs} & Baseline~\cite{ibarz:2022genaralist}  & Baseline + OOD Val & TEAM (Ours)\\
            \midrule
    		Div.\&C.           & ${13.05\%\pm0.14}$     & ${20.30\%\pm0.85}$     & ${65.23\%\pm4.44}$     & $\mathbf{76.36\%\pm1.34}$     & ${65.70\%\pm3.72}$     & ${69.79\%\pm1.60}$\\
    		DP                 & ${67.94\%\pm8.20}$     & ${65.10\%\pm6.44}$     & ${70.58\%\pm6.48}$     & ${81.99\%\pm4.98}$     & $\mathbf{88.83\%\pm5.06}$     & ${83.61\%\pm10.57}$\\
    		Geometry           & ${45.14\%\pm11.95}$    & ${73.11\%\pm17.19}$    & ${61.19\%\pm7.01}$     & $\mathbf{94.09\%\pm2.30}$     & ${88.02\%\pm6.63}$     & ${94.03\%\pm1.57}$\\
            Graphs             & ${24.12\%\pm5.30}$     & ${62.79\%\pm8.75}$     & ${60.25\%\pm8.42}$     & ${81.41\%\pm6.21}$     & ${77.89\%\pm21.49}$     & $\mathbf{81.6\%\pm23.33}$\\
            Greedy             & ${53.42\%\pm20.82}$    & ${82.39\%\pm3.01}$     & ${75.84\%\pm6.59}$     & ${91.21\%\pm2.95}$     & ${90.08\%\pm5.06}$     & $\mathbf{91.80\%\pm4.68}$\\
            Search             & ${34.35\%\pm21.67}$    & ${41.20\%\pm19.87}$    & ${56.11\%\pm21.56}$    & ${58.61\%\pm24.3}$     & ${61.89\%\pm45.48}$     & $\mathbf{62.79\%\pm43.66}$\\
            Sorting            & ${71.53\%\pm1.41}$     & ${11.83\%\pm2.78}$     & ${15.45\%\pm8.46}$     & ${60.37\%\pm12.7}$     & $\mathbf{72.08\%\pm21.31}$     & ${68.75\%\pm18.64}$\\
            Strings            & ${1.51\%\pm0.46}$      & ${3.21\%\pm0.94}$      & ${2.04\%\pm0.20}$      & ${49.09\%\pm23.5}$     & ${75.33\%\pm22.57}$     & $\mathbf{81.24\%\pm25.05}$\\
            \midrule
            Average            & ${38.88\%}$            & ${44.99\%}$            & ${50.84\%}$            & ${74.14\%}$            & ${77.65\%}$            & $\mathbf{79.23\%}$\\
            Rank Avg.          & $5.38$                 & $4.75$                 & $4.63$                 & $2.38$                 & $2.25$                 & $\mathbf{1.63}$\\
            \midrule
            $>90\%$            & ${0/30}$               & ${6/30}$               & ${3/30}$               & ${11/30}$              & ${11/30}$              & $\mathbf{15/30}$\\
            $>50\%$            & ${10/30}$              & ${17/30}$              & ${18/30}$              & ${24/30}$              & ${26/30}$              & $\mathbf{27/30}$\\
		\bottomrule
	\end{tabular}}
\end{table}

\section{Conclusion}
In this work, we present a new \textbf{TEA} (\textbf{T}riplet \textbf{E}dge \textbf{A}ttention) method, designed to compute edge latent during the algorithmic reasoning process. Based on the method, we propose the \textbf{TEAM} (\textbf{T}riplet \textbf{E}dge \textbf{A}ttention \textbf{M}essage Passing Neural Network) model. Our TEAM model with OOD validation set exhibited improvements across the CLRS benchmark algorithms and demonstrated a particularly significant enhancement with string algorithms, as compared to the baseline model \cite{ibarz:2022genaralist}. In conclusion, we have developed a novel model and training methodology for algorithmic reasoning tasks and we are confident that additional developments within the reasoning field can lead even to higher performance levels, eventually unlocking the potential for a generally effective model.  

\bibliographystyle{unsrtnat}
\bibliography{reference}

\appendix
\section{Implementation details}


\subsection{Processor}
\begin{figure}[h]
	\centering
	\includegraphics[width=01\linewidth]{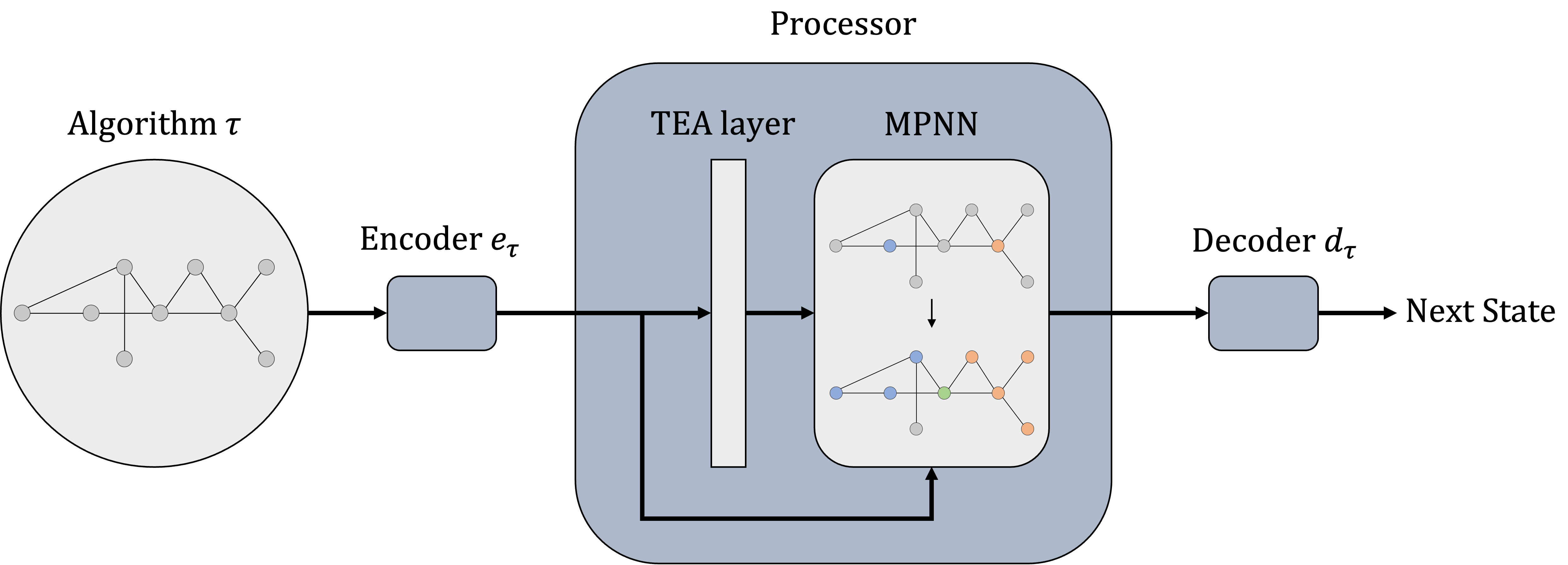}
	\caption{The \textbf{TEAM} Processor Network composed of Triplet Edge Attention Layer and Message Passing Neural Network. }
	\label{fig:TEAM_network}
\end{figure}

The processor is the key component in the encoder-processor-decoder architecture for algorithmic reasoning. We designed a new \textbf{TEAM} (\textbf{T}riplet \textbf{E}dge \textbf{A}ttention \textbf{M}essage Passing Neural Network) network for training algorithms, combining the triplet edge attention (TEA) layer to a fully connected MPNN (Message Passing Neural Network) \cite{gilmerz:2017mpnn} as Figure \ref{fig:TEAM_network}. 

The embedded results from the encoder are processed through the TEA layer to compute the edge latents. Afterward, the computed edge latents and the embeddings from the encoder are given as input for the fully connected MPNN. The MPNN is responsible for predicting the algorithm's hints, which are then fed to the decoder to get the current state. In contrast to the previous triplet reasoning approach \citep{ibarz:2022genaralist}, which fed the edge latent directly to the decoder, we have replaced the edge feature to our TEA-derived edge latent for MPNN computation. This change permits reasoning on nodes utilizing our edge latents. 

\subsection{Encoder and Decoder}
The encoder and decoder from the baseline were adopted in our approach. For an algorithm $\tau$, encoder $\mathbf{f_\tau}$ and decoder $\mathbf{g_\tau}$ are defined tasks specifically since each algorithm requires different hints to predict the algorithm's subsequent state. The encoder performs linear embeddings of node features, edge features, graph features, and the hints of the preceding state into a higher-dimensional space. The embedded results are then fed to our processor network to predict the hints for the next stage of the algorithm. Finally, the decoder estimates the algorithmic reasoning for the current state by linearly decoding the processor network's output. 


\subsection{Additional Experiment Results}


\begin{figure}[h]
    \begin{subfigure}[b]{0.245\linewidth}
        \centering
        \includegraphics[width=\textwidth]{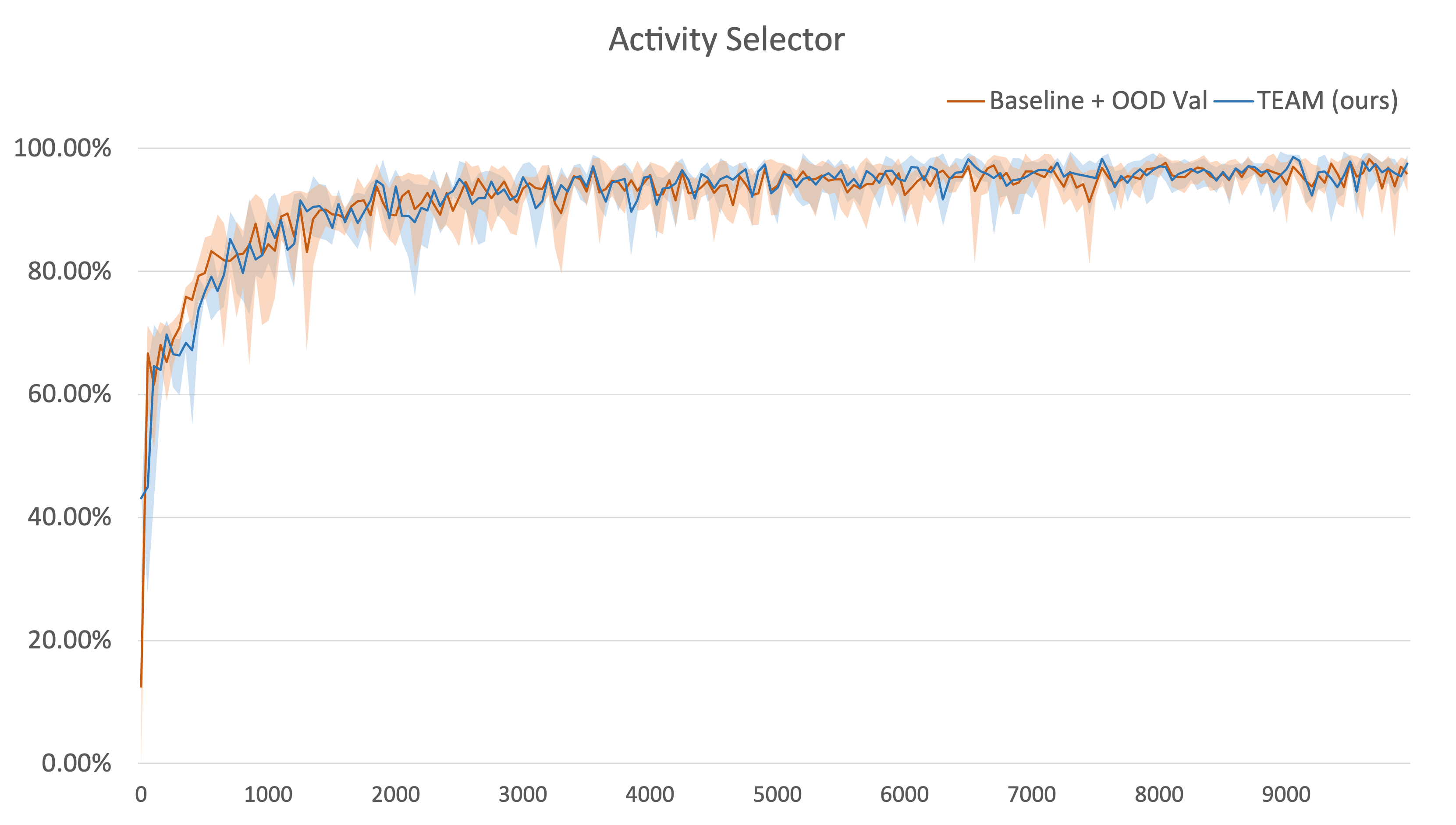}
    \end{subfigure}
    \begin{subfigure}[b]{0.245\linewidth}
        \centering
        \includegraphics[width=\textwidth]{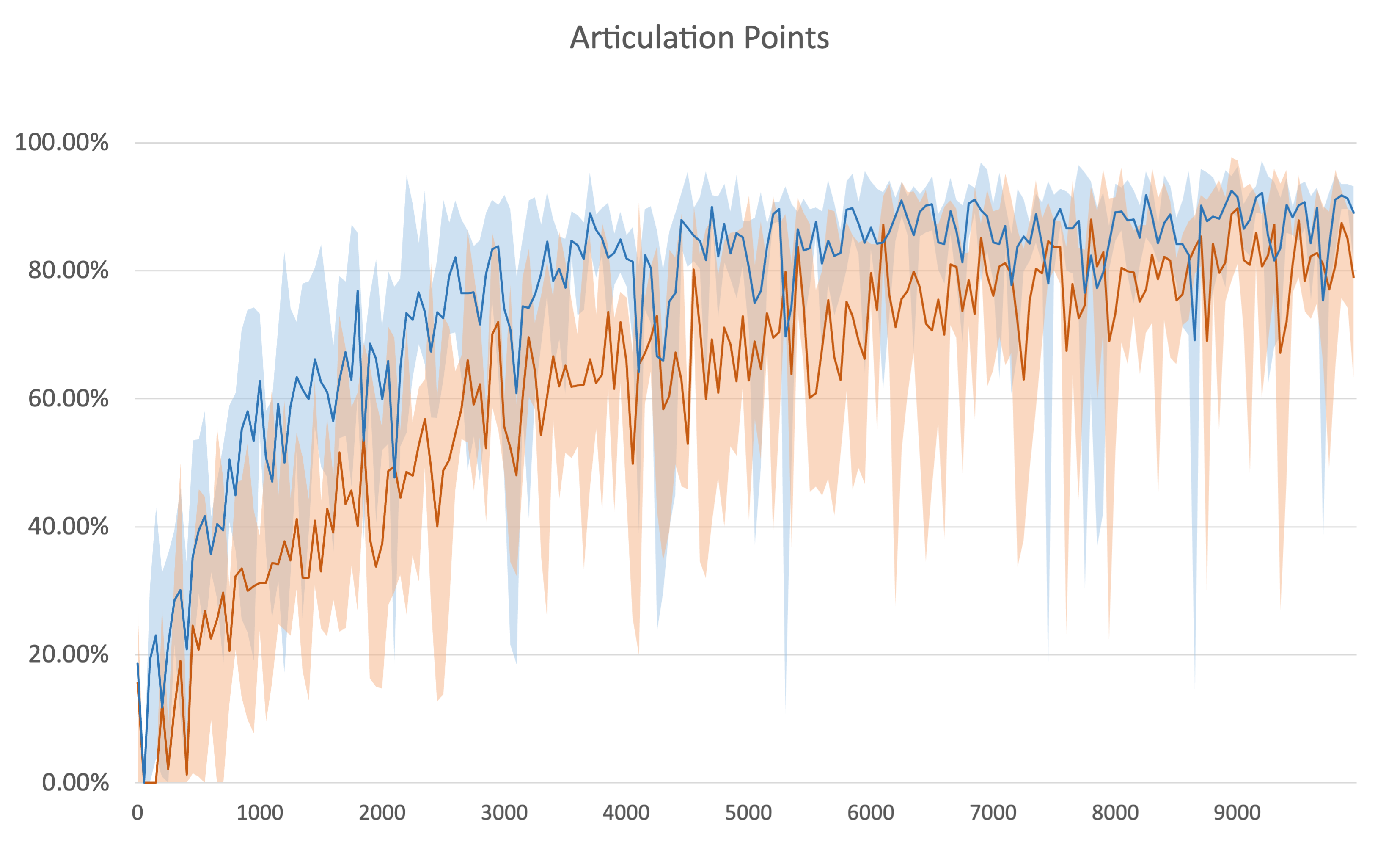}
    \end{subfigure}
    \begin{subfigure}[b]{0.245\linewidth}
        \centering
        \includegraphics[width=\textwidth]{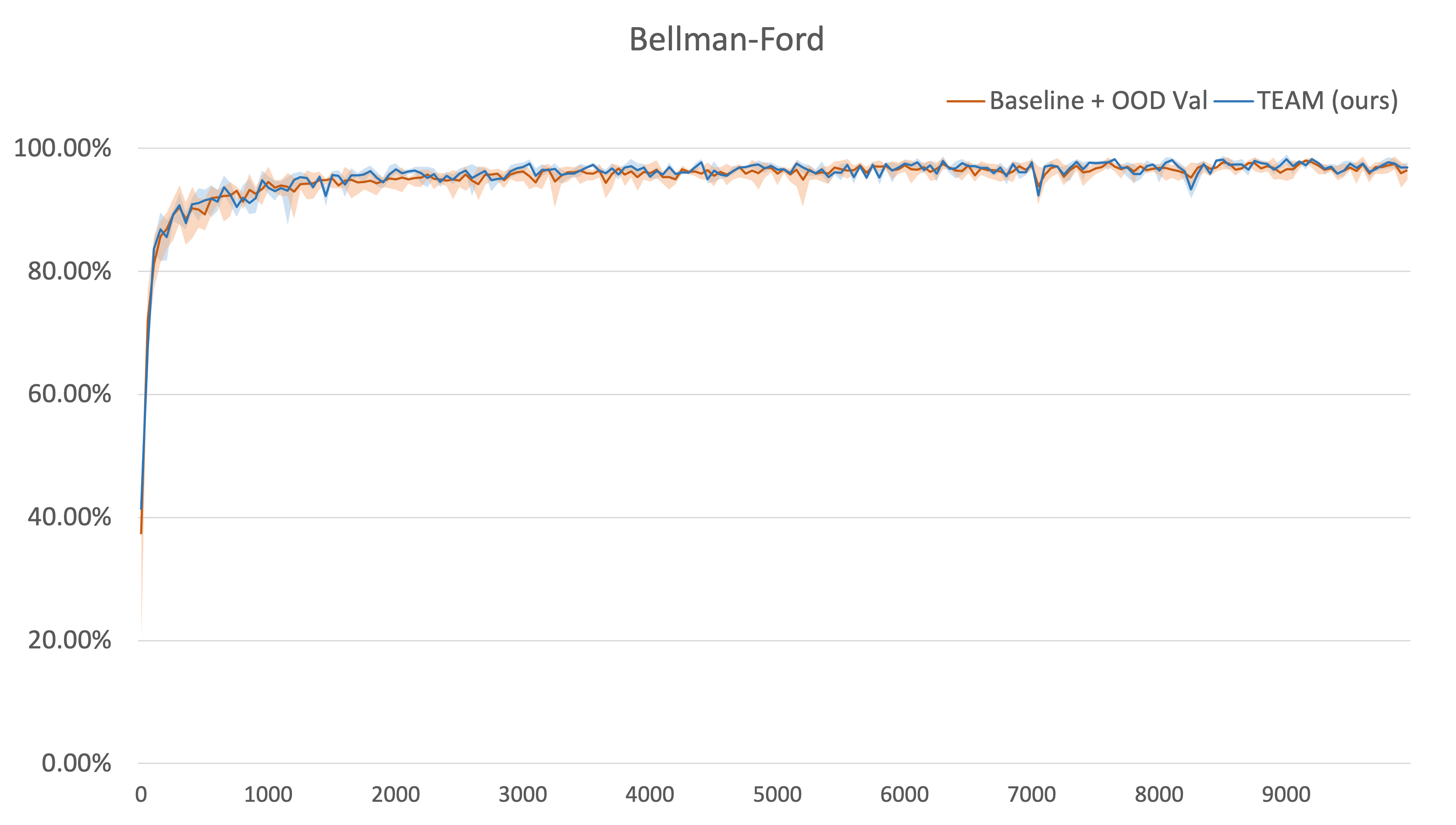}
    \end{subfigure}
    \begin{subfigure}[b]{0.245\linewidth}
        \centering
        \includegraphics[width=\textwidth]{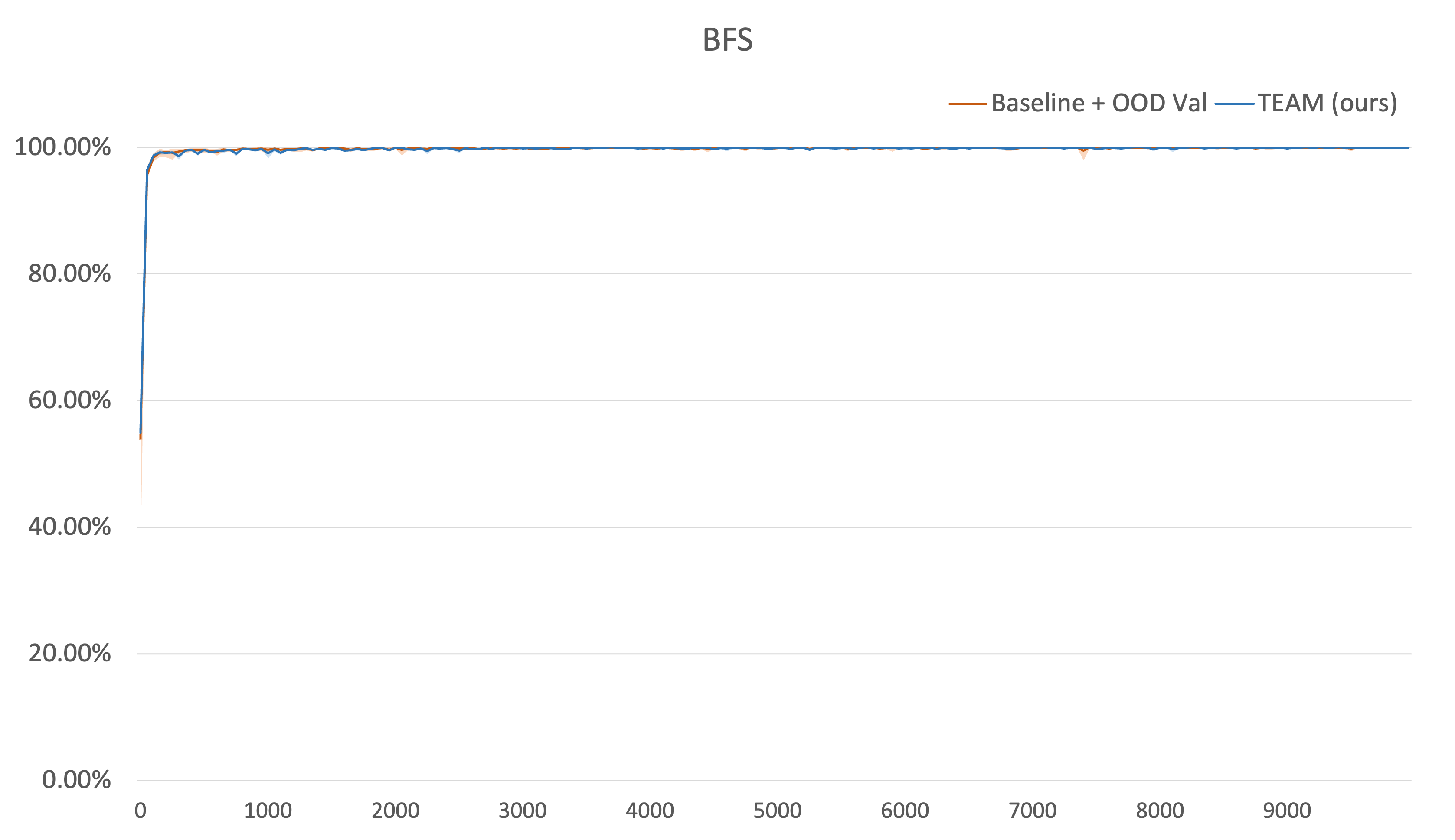}
    \end{subfigure}
    \begin{subfigure}[b]{0.245\linewidth}
        \centering
        \includegraphics[width=\textwidth]{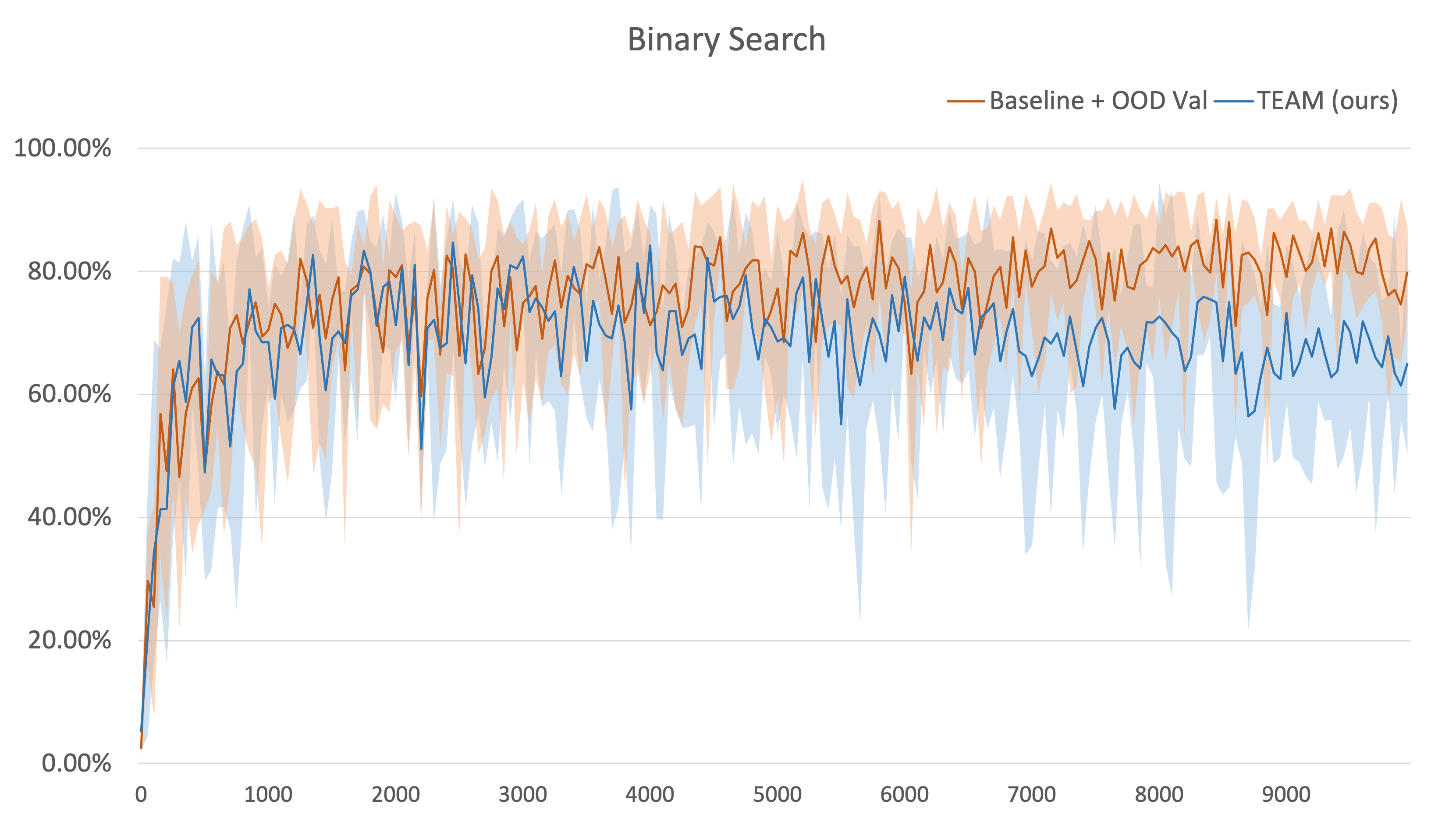}
    \end{subfigure}
    \begin{subfigure}[b]{0.245\linewidth}
        \centering
        \includegraphics[width=\textwidth]{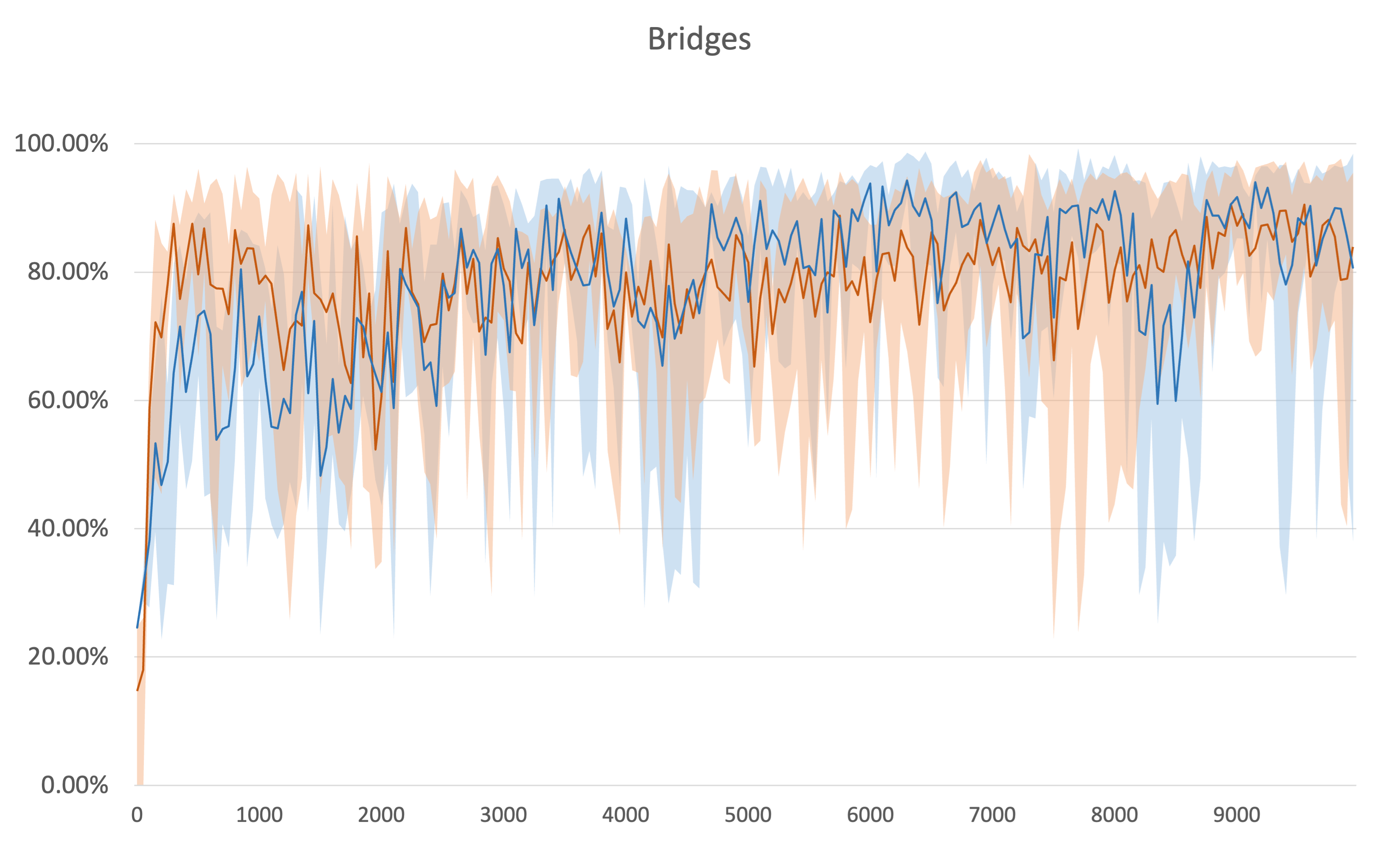}
    \end{subfigure}
    \begin{subfigure}[b]{0.245\linewidth}
        \centering
        \includegraphics[width=\textwidth]{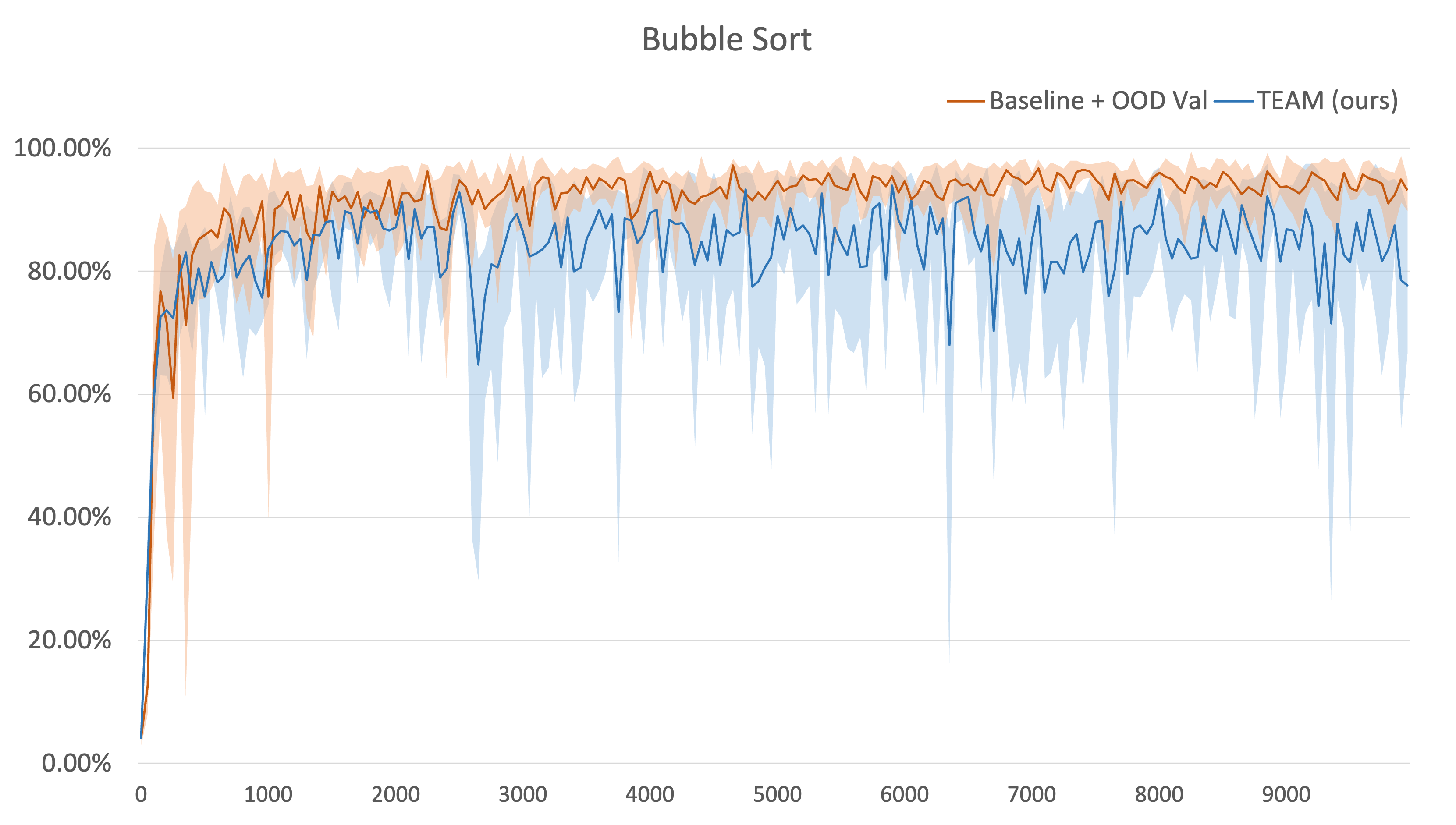}
    \end{subfigure}
    \begin{subfigure}[b]{0.245\linewidth}
        \centering
        \includegraphics[width=\textwidth]{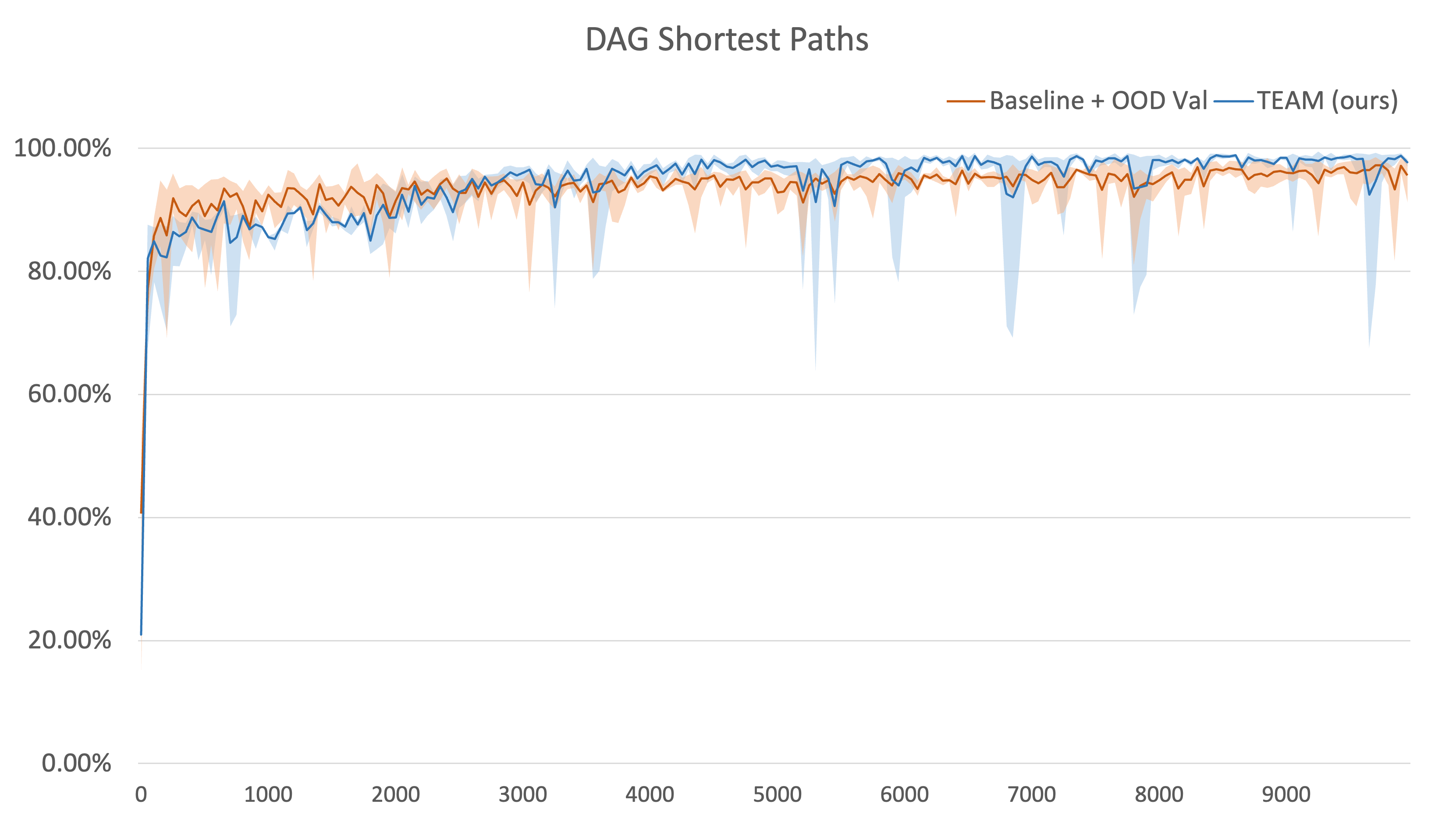}
    \end{subfigure}
    \begin{subfigure}[b]{0.245\linewidth}
        \centering
        \includegraphics[width=\textwidth]{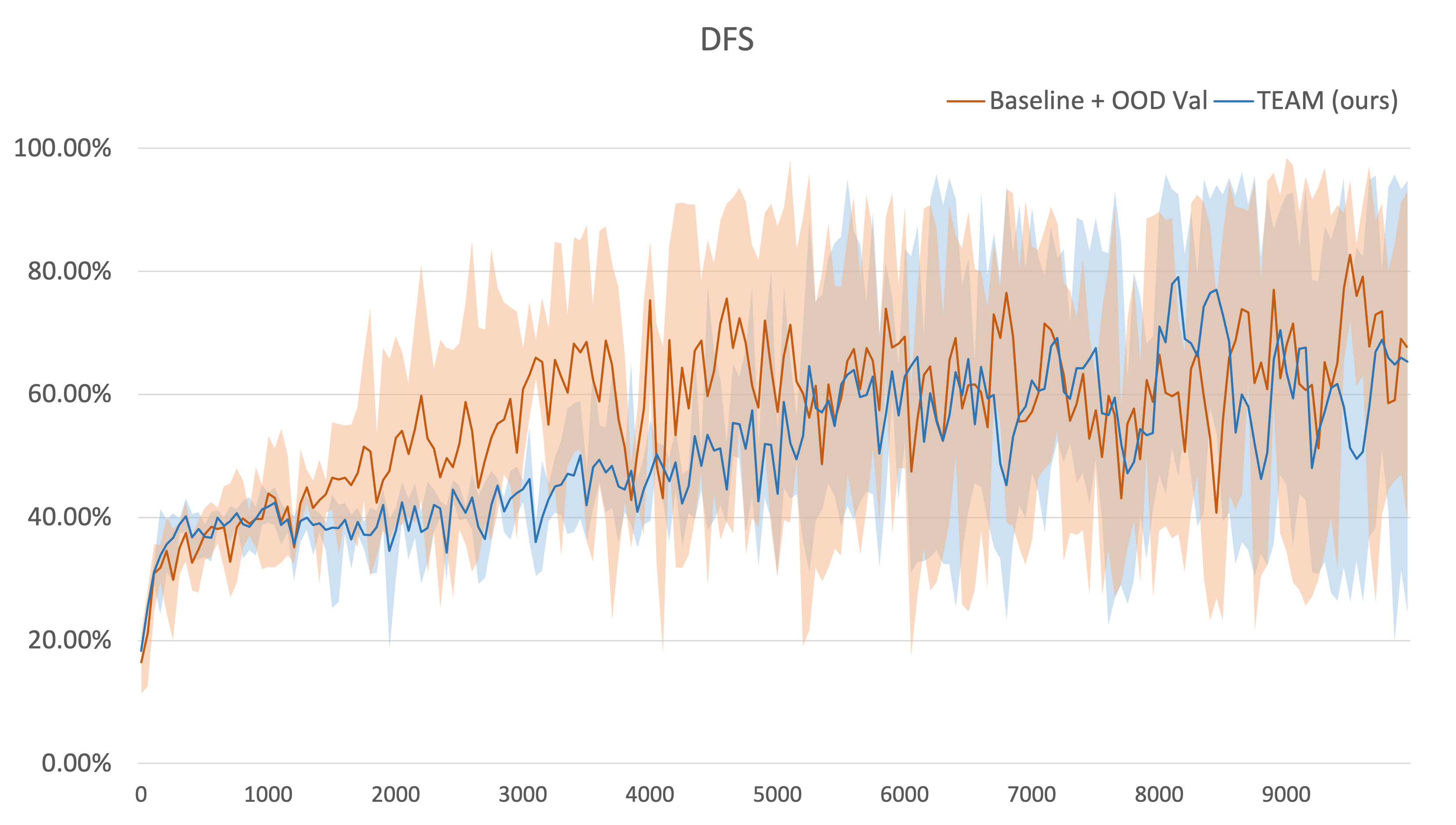}
    \end{subfigure}
    \begin{subfigure}[b]{0.245\linewidth}
        \centering
        \includegraphics[width=\textwidth]{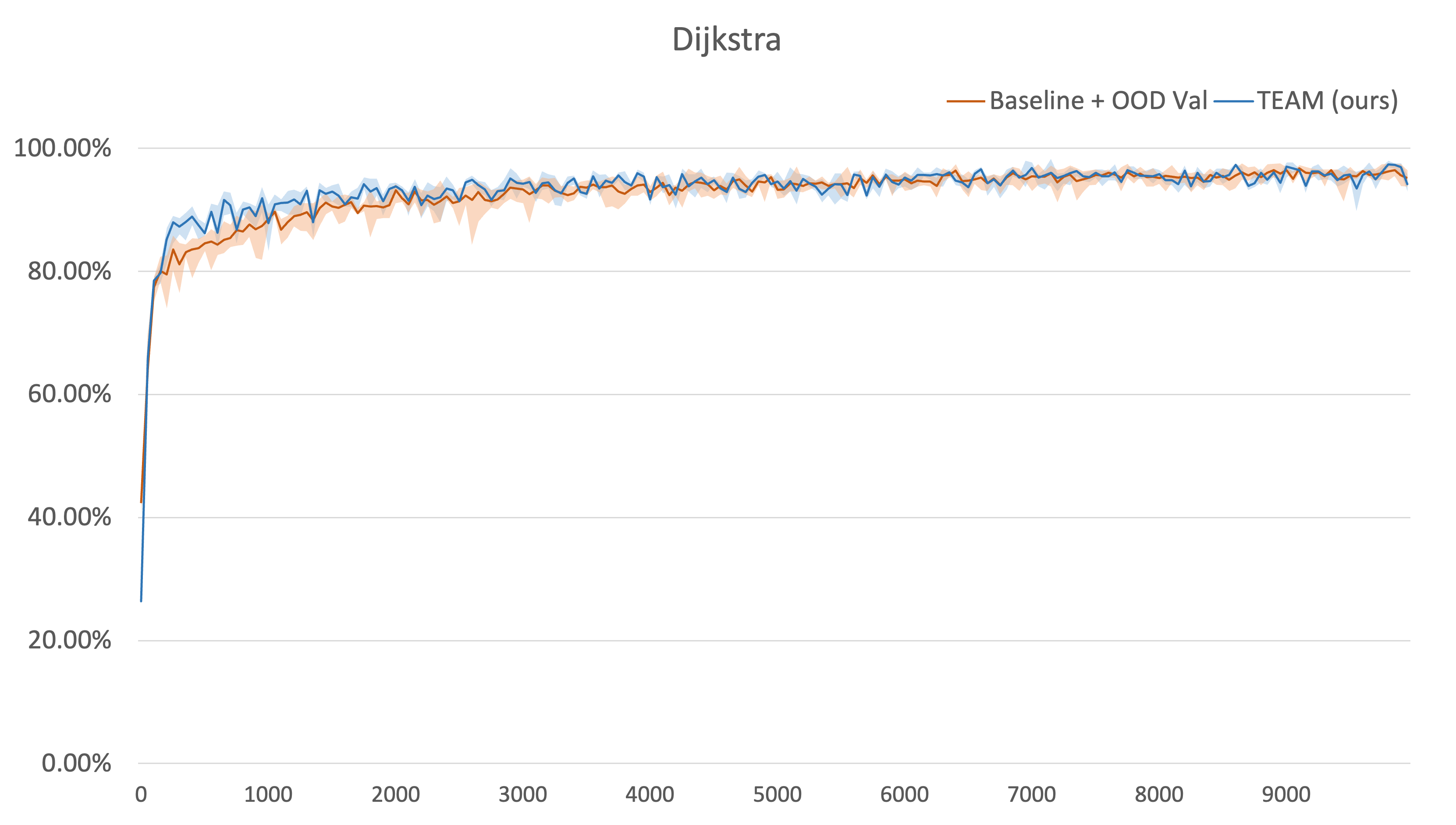}
    \end{subfigure}
    \begin{subfigure}[b]{0.245\linewidth}
        \centering
        \includegraphics[width=\textwidth]{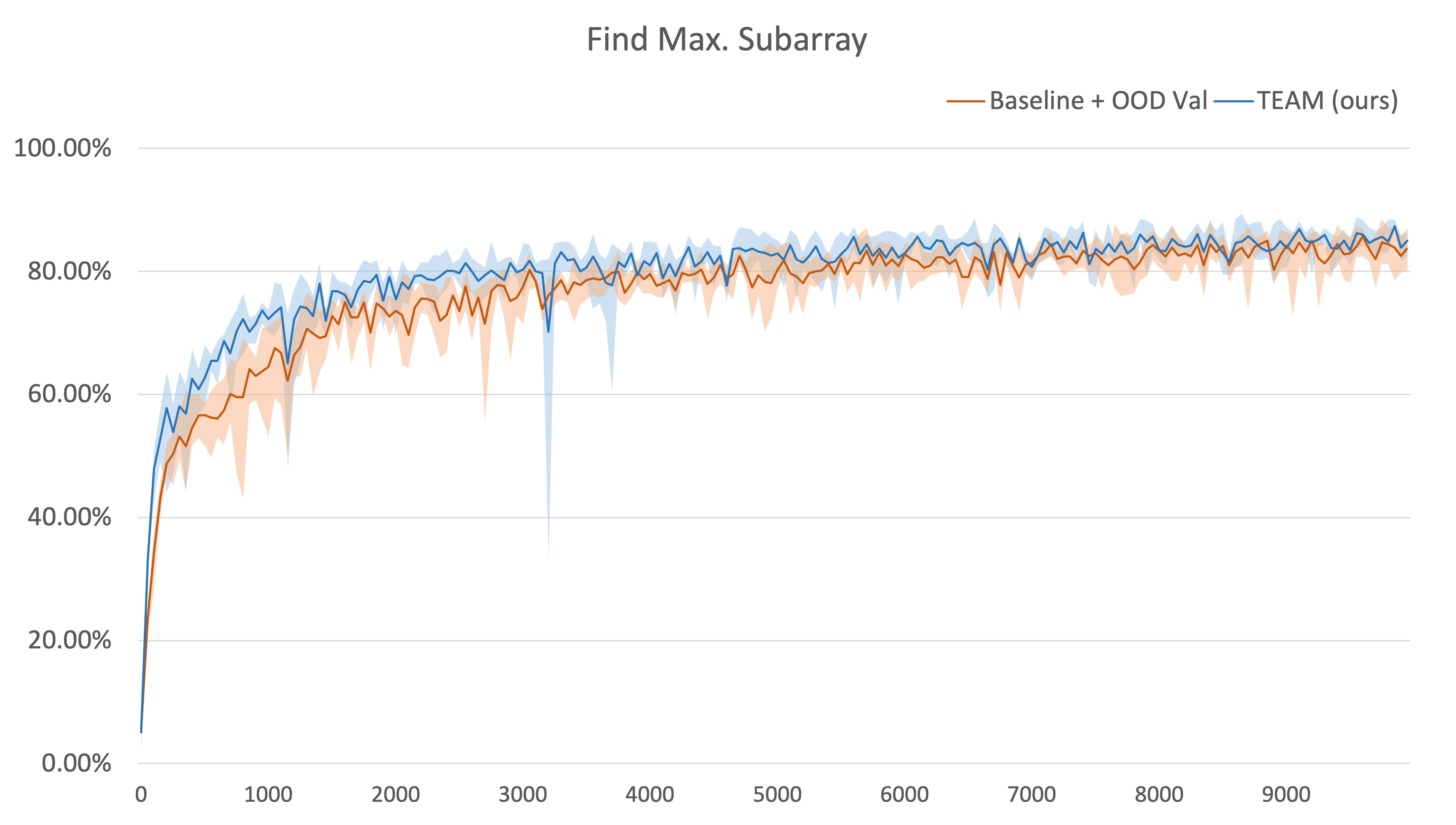}
    \end{subfigure}
    \begin{subfigure}[b]{0.245\linewidth}
        \centering
        \includegraphics[width=\textwidth]{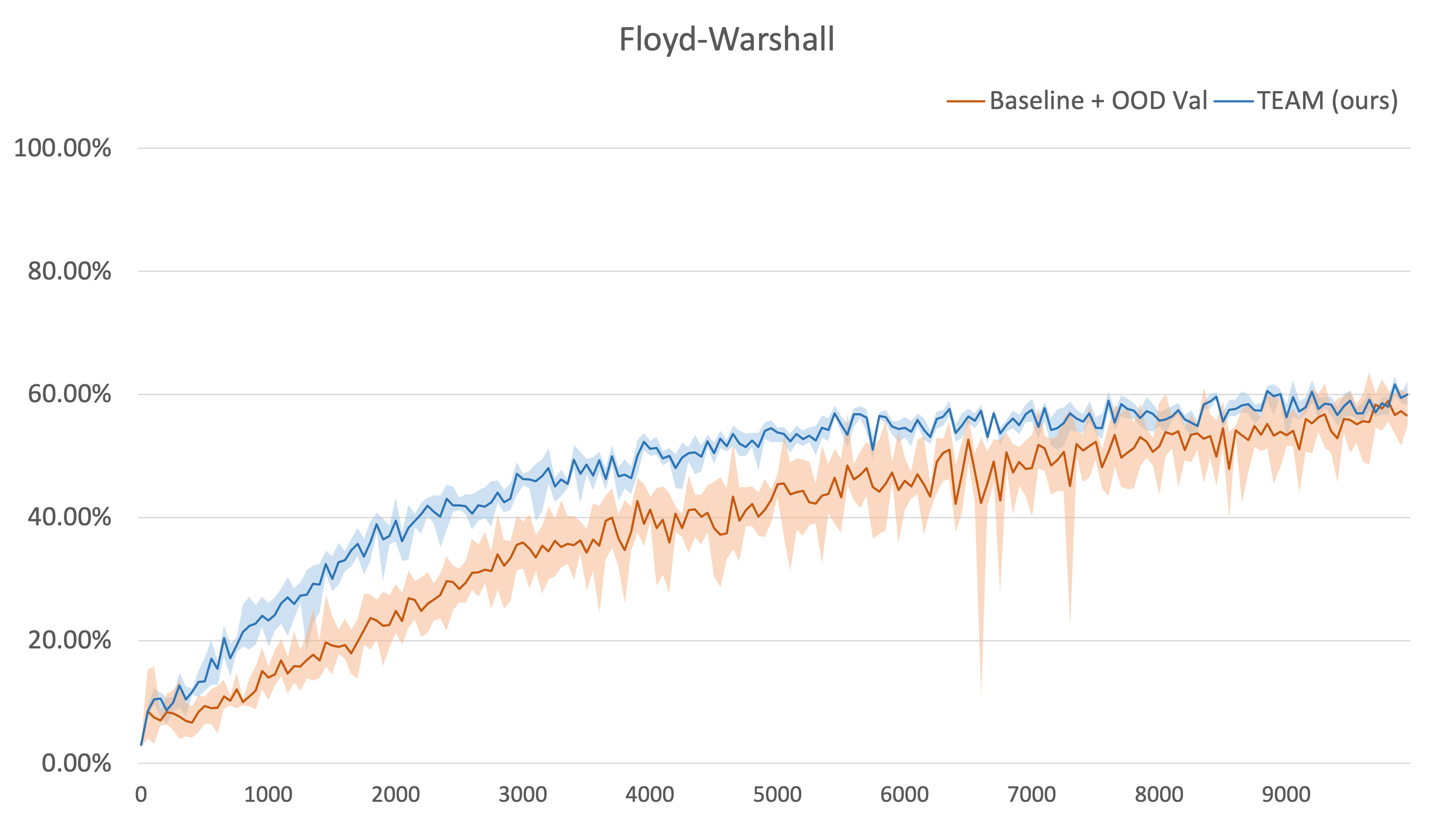}
    \end{subfigure}
    \begin{subfigure}[b]{0.245\linewidth}
        \centering
        \includegraphics[width=\textwidth]{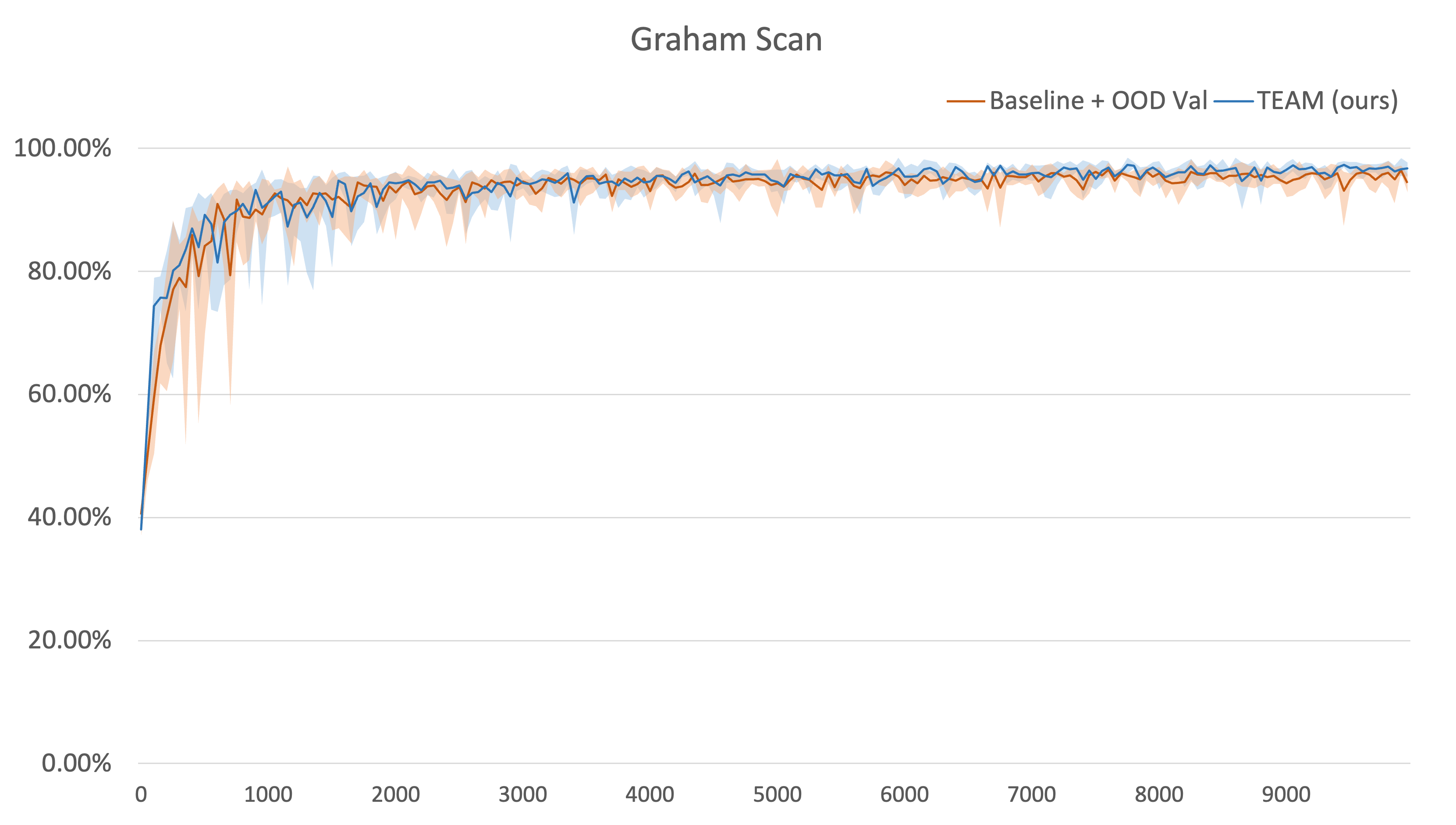}
    \end{subfigure}
    \begin{subfigure}[b]{0.245\linewidth}
        \centering
        \includegraphics[width=\textwidth]{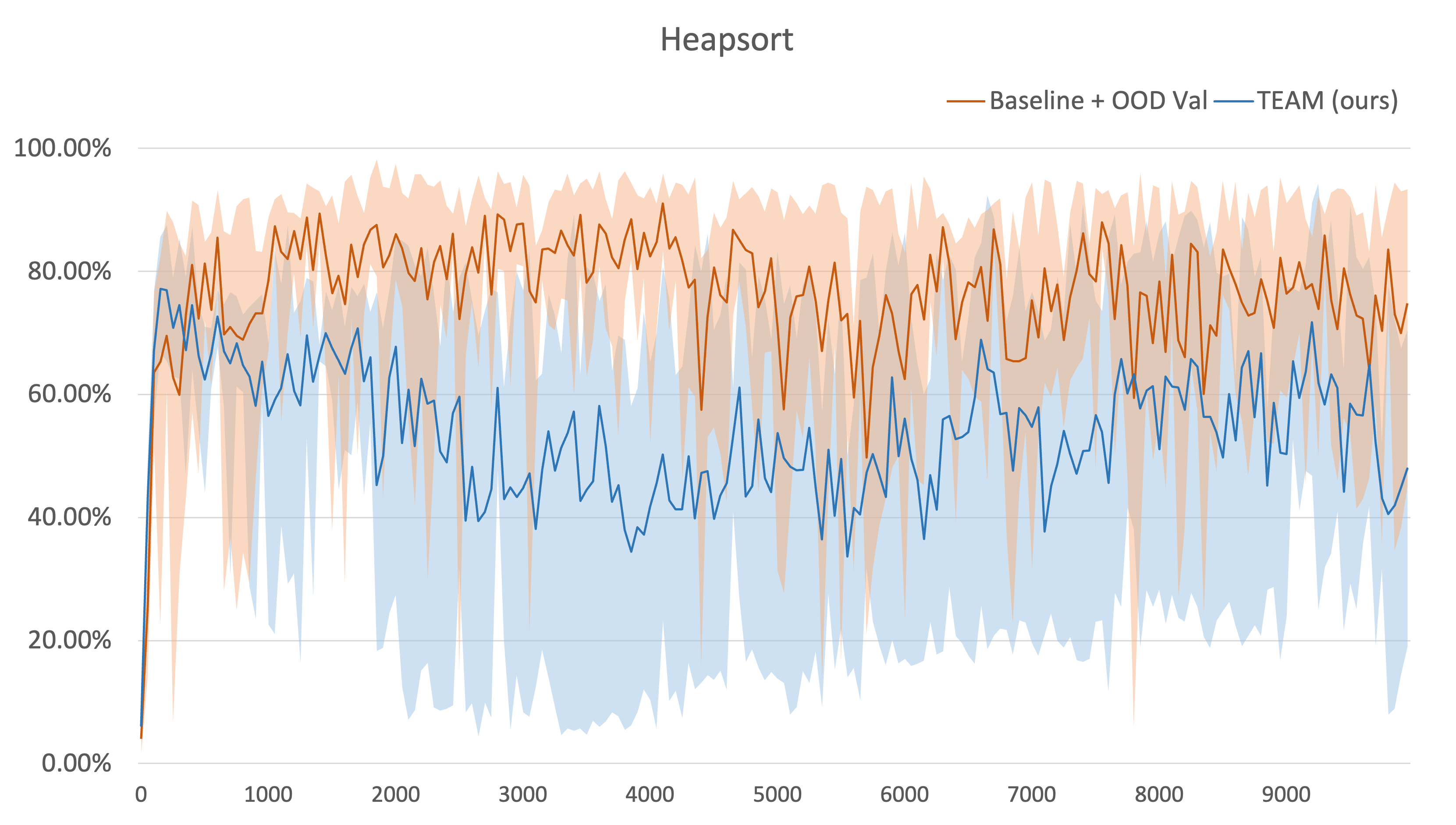}
    \end{subfigure}
    \begin{subfigure}[b]{0.245\linewidth}
        \centering
        \includegraphics[width=\textwidth]{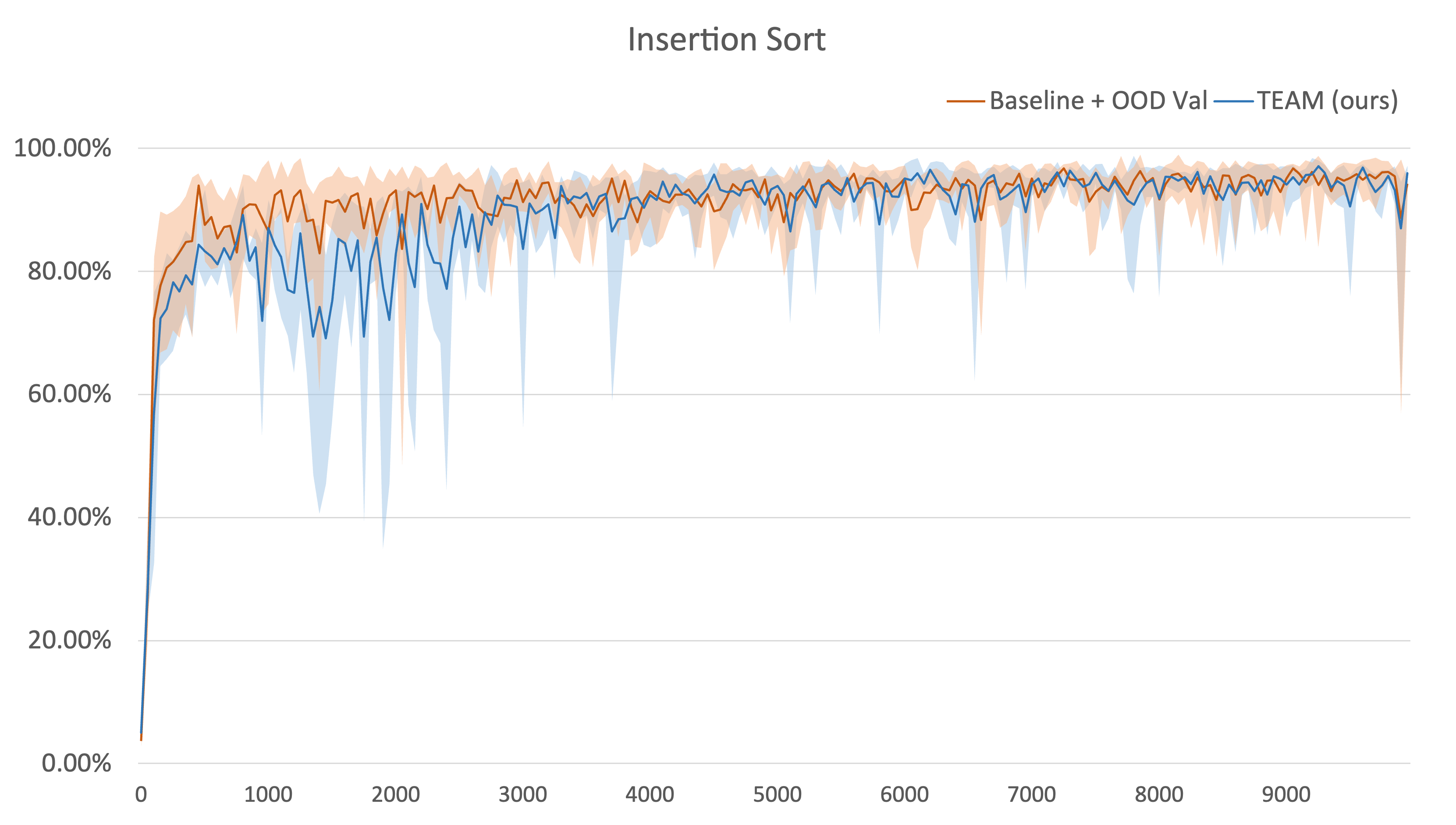}
    \end{subfigure}
    \begin{subfigure}[b]{0.245\linewidth}
        \centering
        \includegraphics[width=\textwidth]{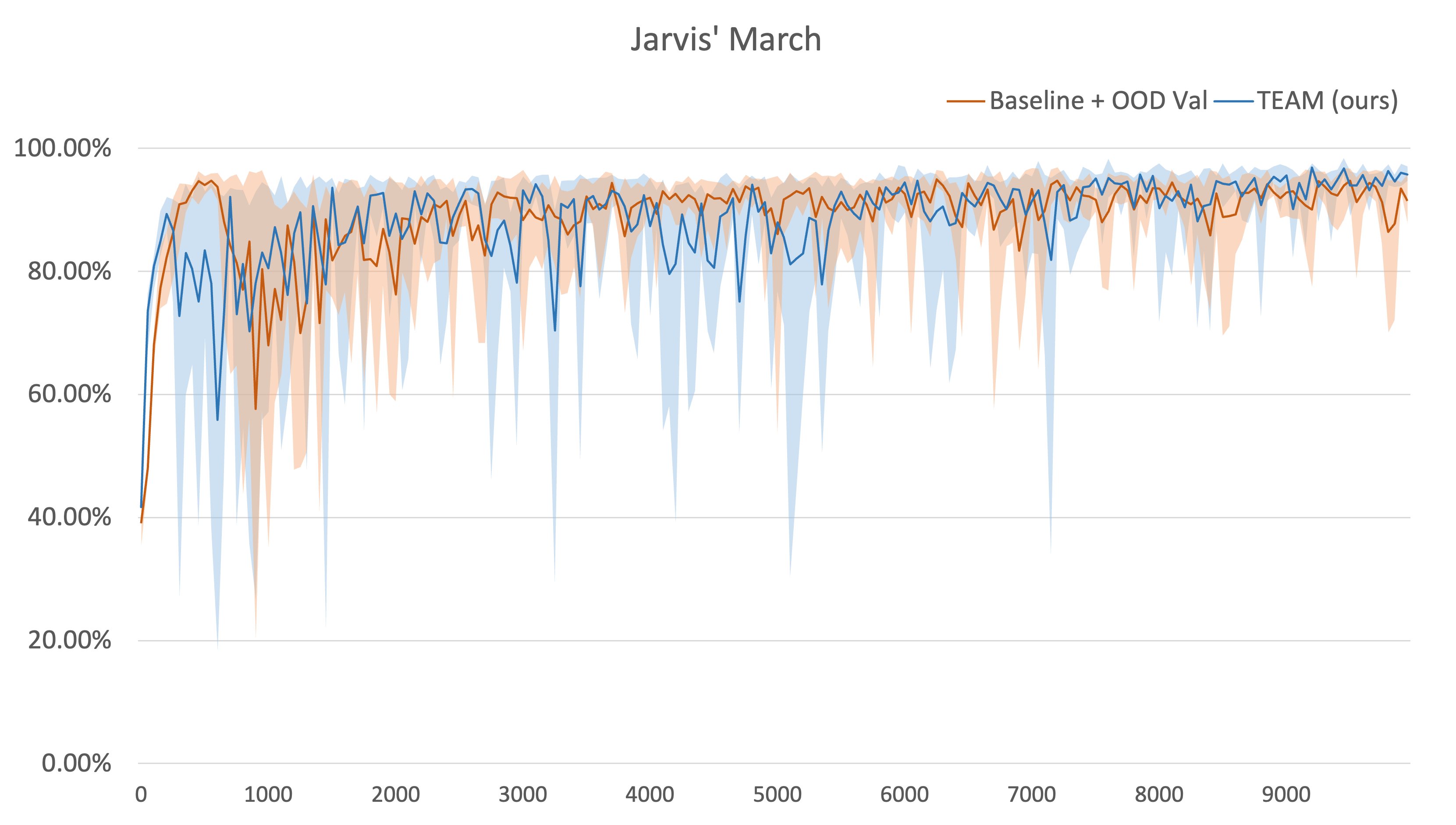}
    \end{subfigure}
    \begin{subfigure}[b]{0.245\linewidth}
        \centering
        \includegraphics[width=\textwidth]{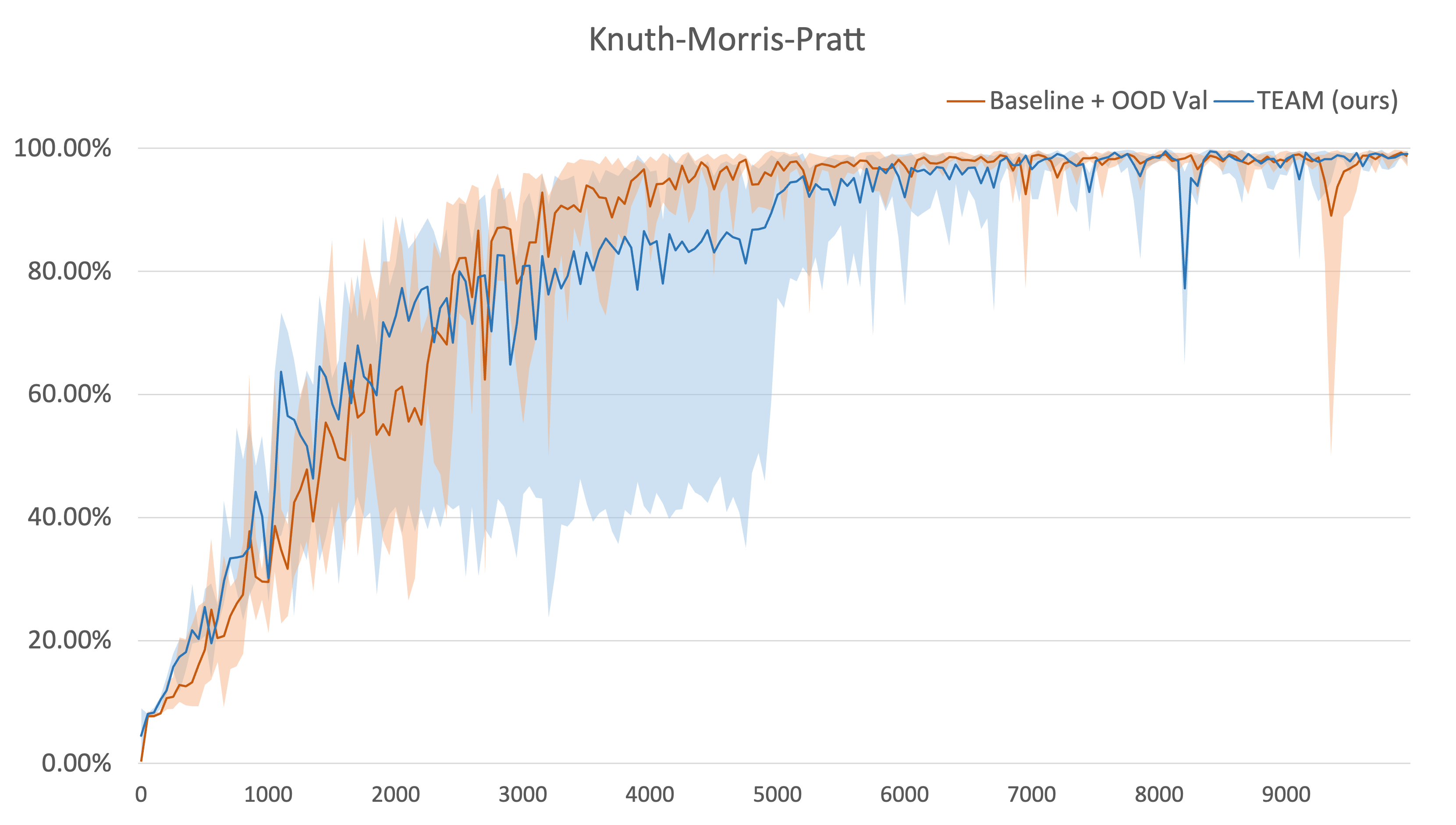}
    \end{subfigure}
    \begin{subfigure}[b]{0.245\linewidth}
        \centering
        \includegraphics[width=\textwidth]{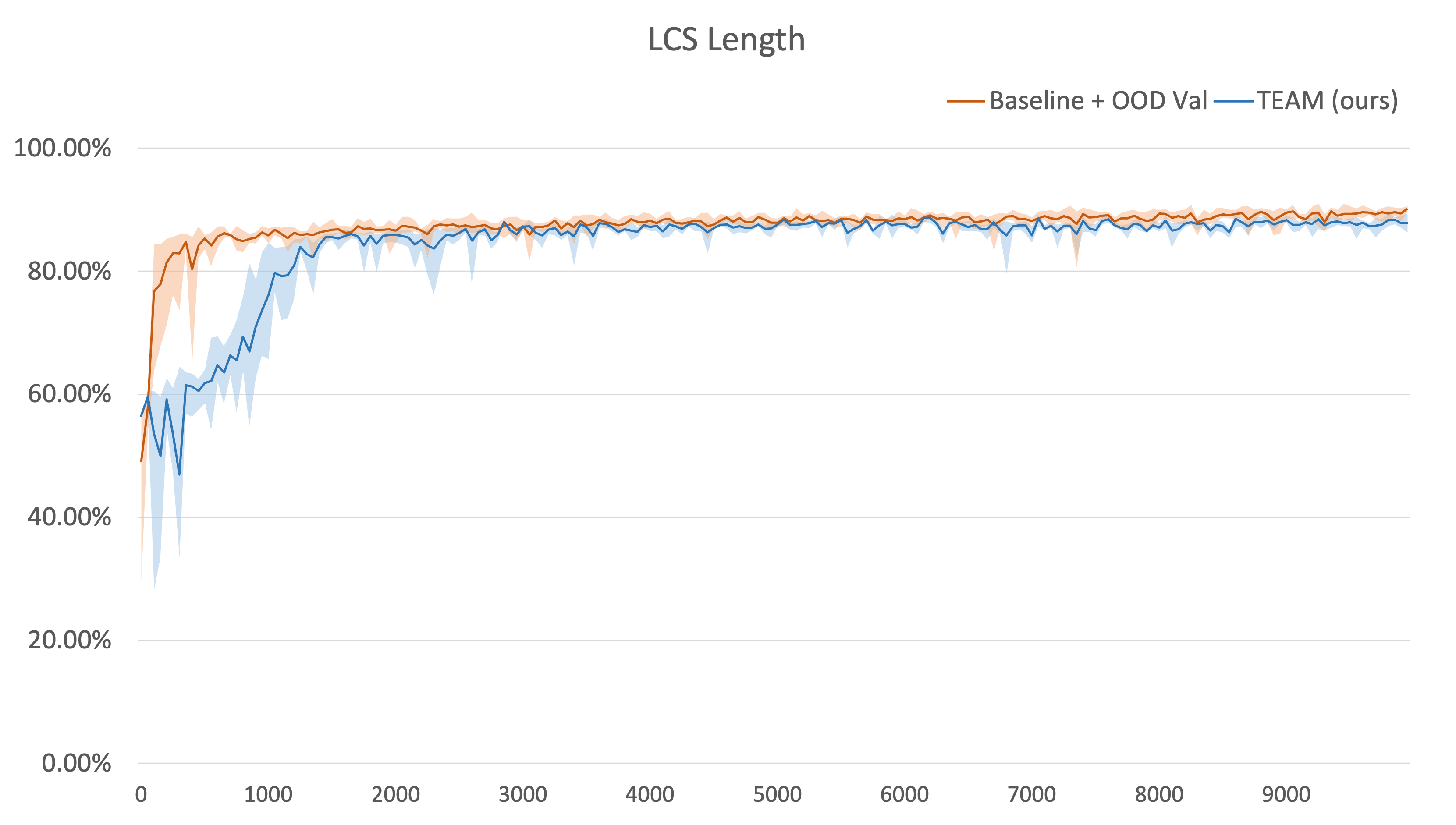}
    \end{subfigure}
    \begin{subfigure}[b]{0.245\linewidth}
        \centering
        \includegraphics[width=\textwidth]{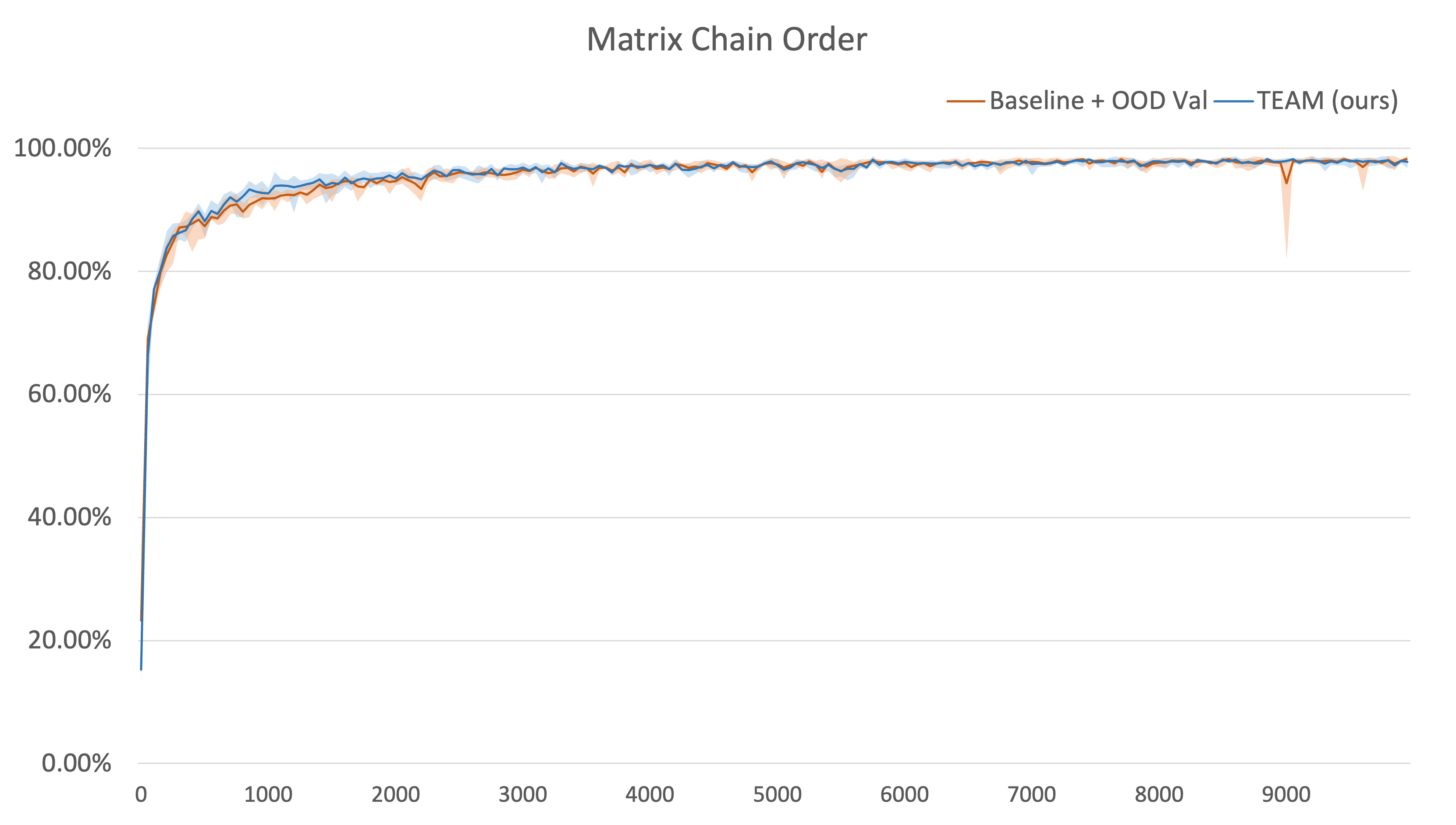}
    \end{subfigure}
    \begin{subfigure}[b]{0.245\linewidth}
        \centering
        \includegraphics[width=\textwidth]{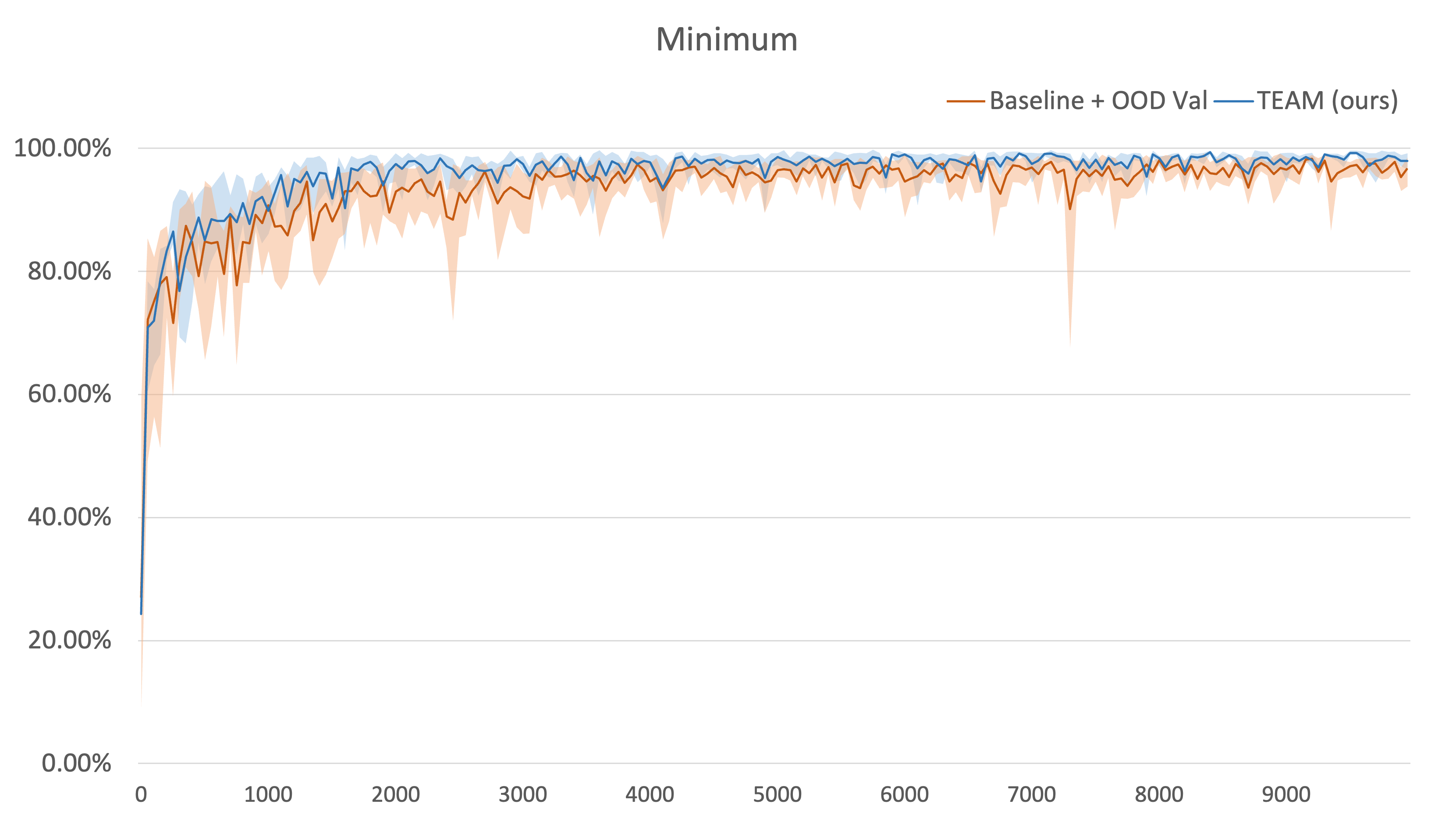}
    \end{subfigure}
    \begin{subfigure}[b]{0.245\linewidth}
        \centering
        \includegraphics[width=\textwidth]{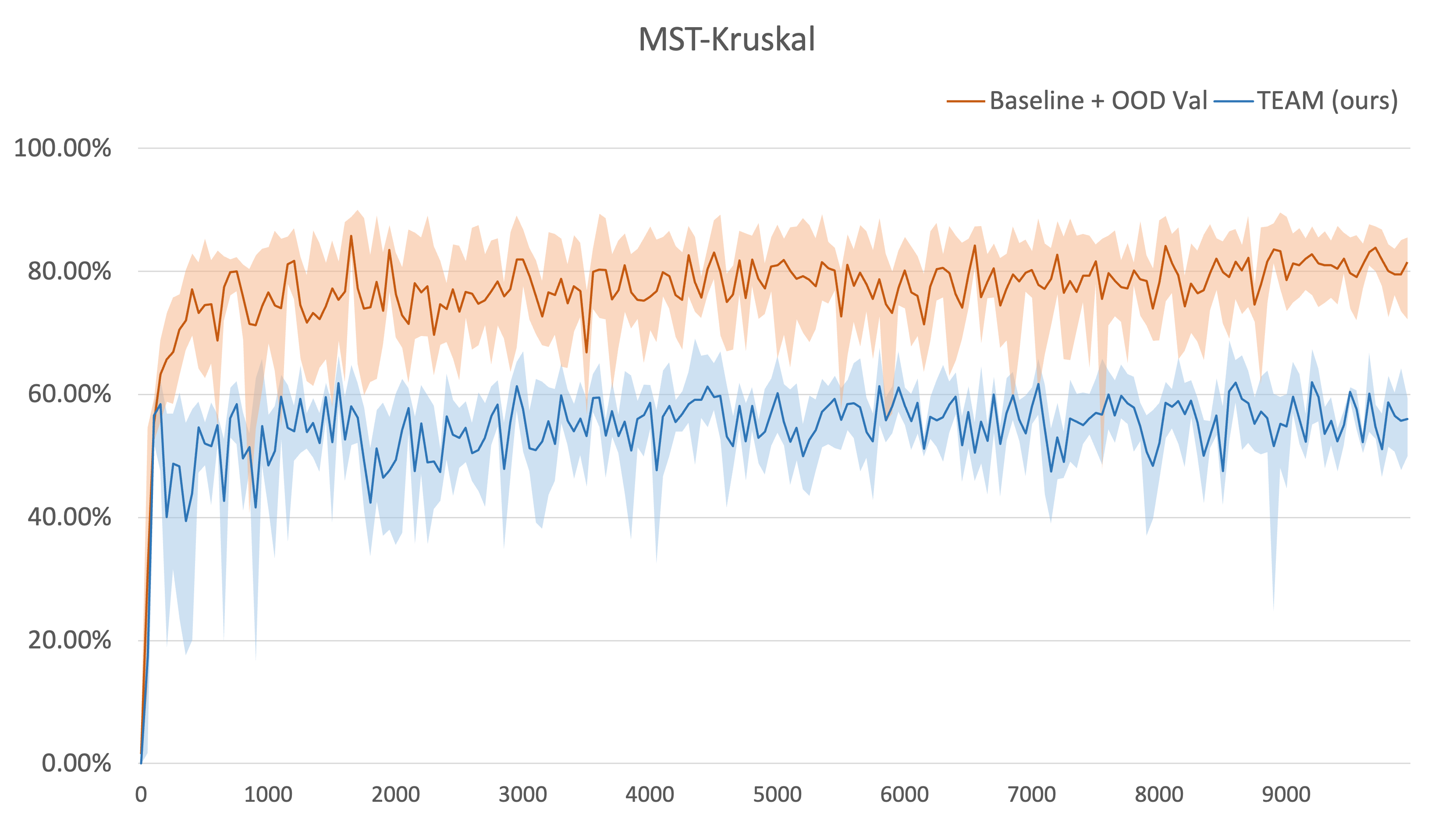}
    \end{subfigure}
    \begin{subfigure}[b]{0.245\linewidth}
        \centering
        \includegraphics[width=\textwidth]{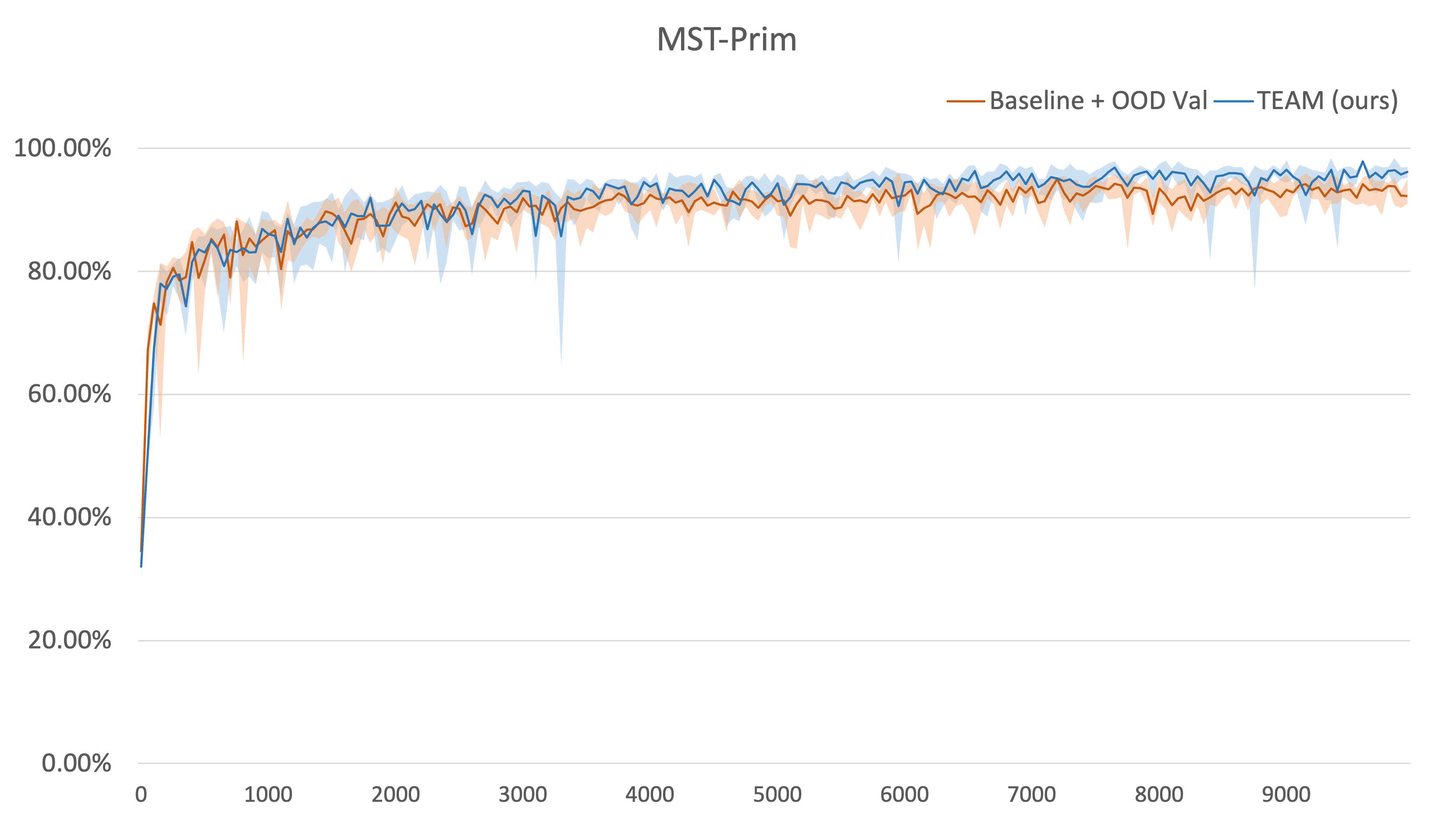}
    \end{subfigure}
    \begin{subfigure}[b]{0.245\linewidth}
        \centering
        \includegraphics[width=\textwidth]{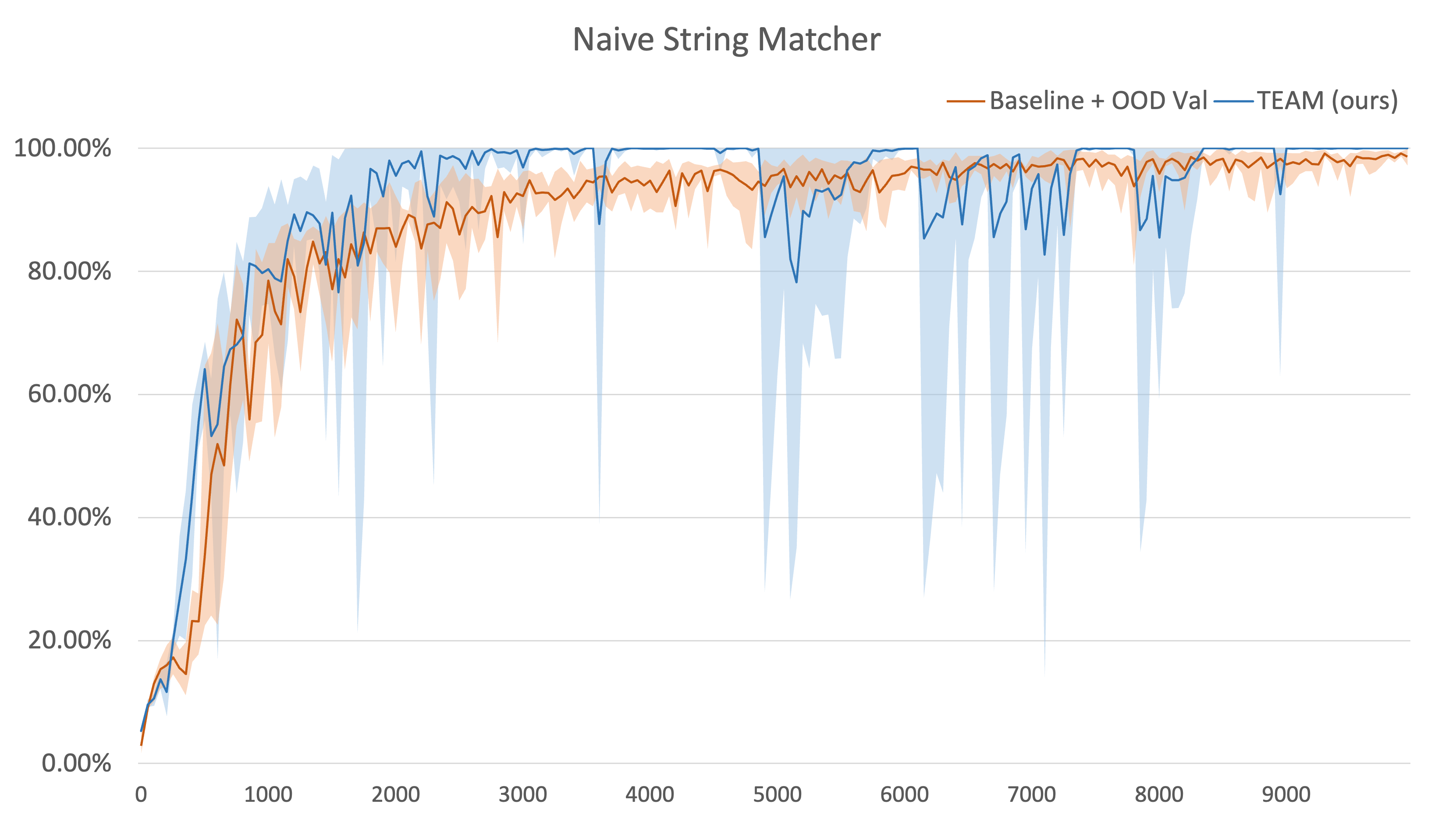}
    \end{subfigure}
    \begin{subfigure}[b]{0.245\linewidth}
        \centering
        \includegraphics[width=\textwidth]{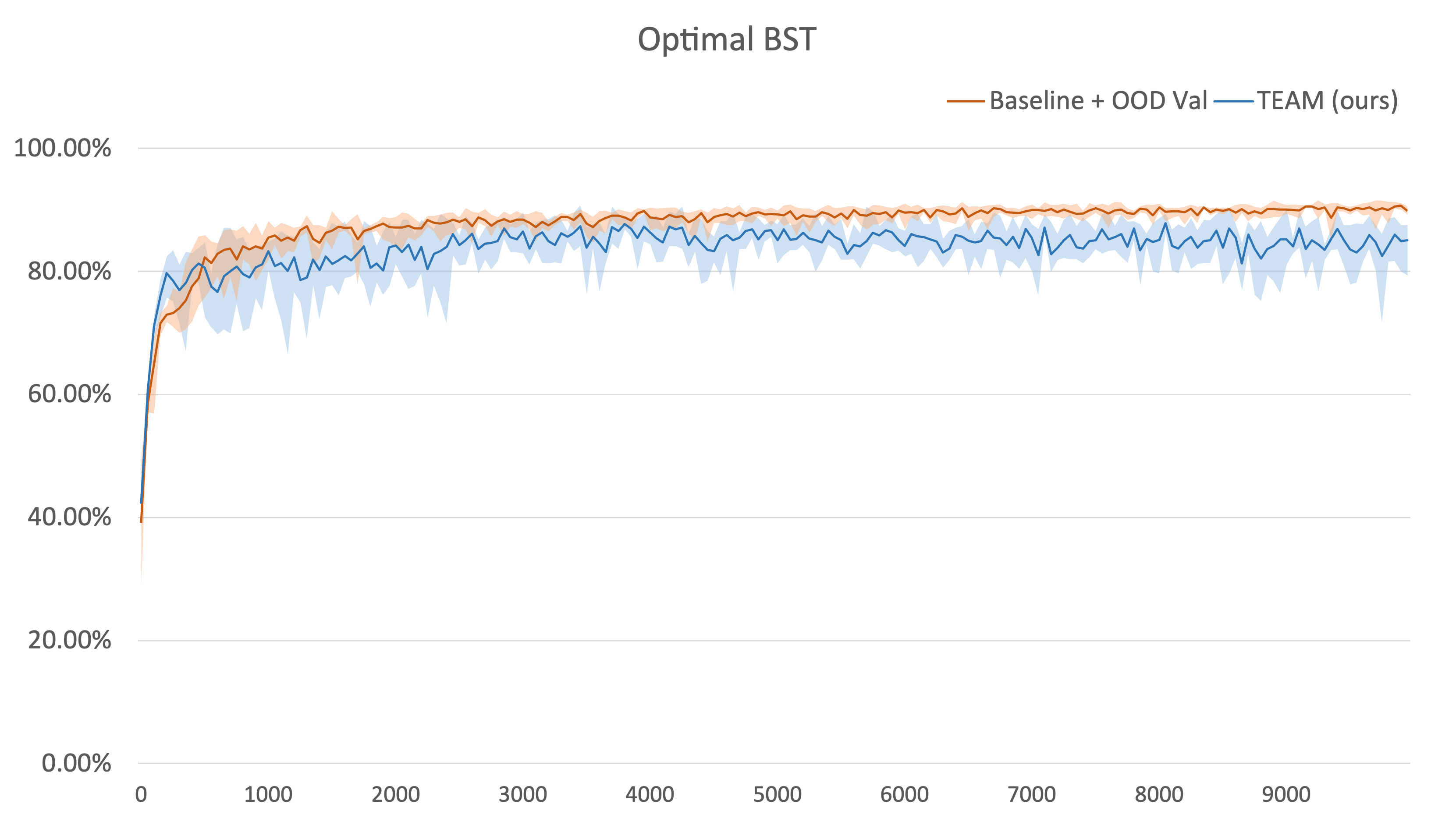}
    \end{subfigure}
    \begin{subfigure}[b]{0.245\linewidth}
        \centering
        \includegraphics[width=\textwidth]{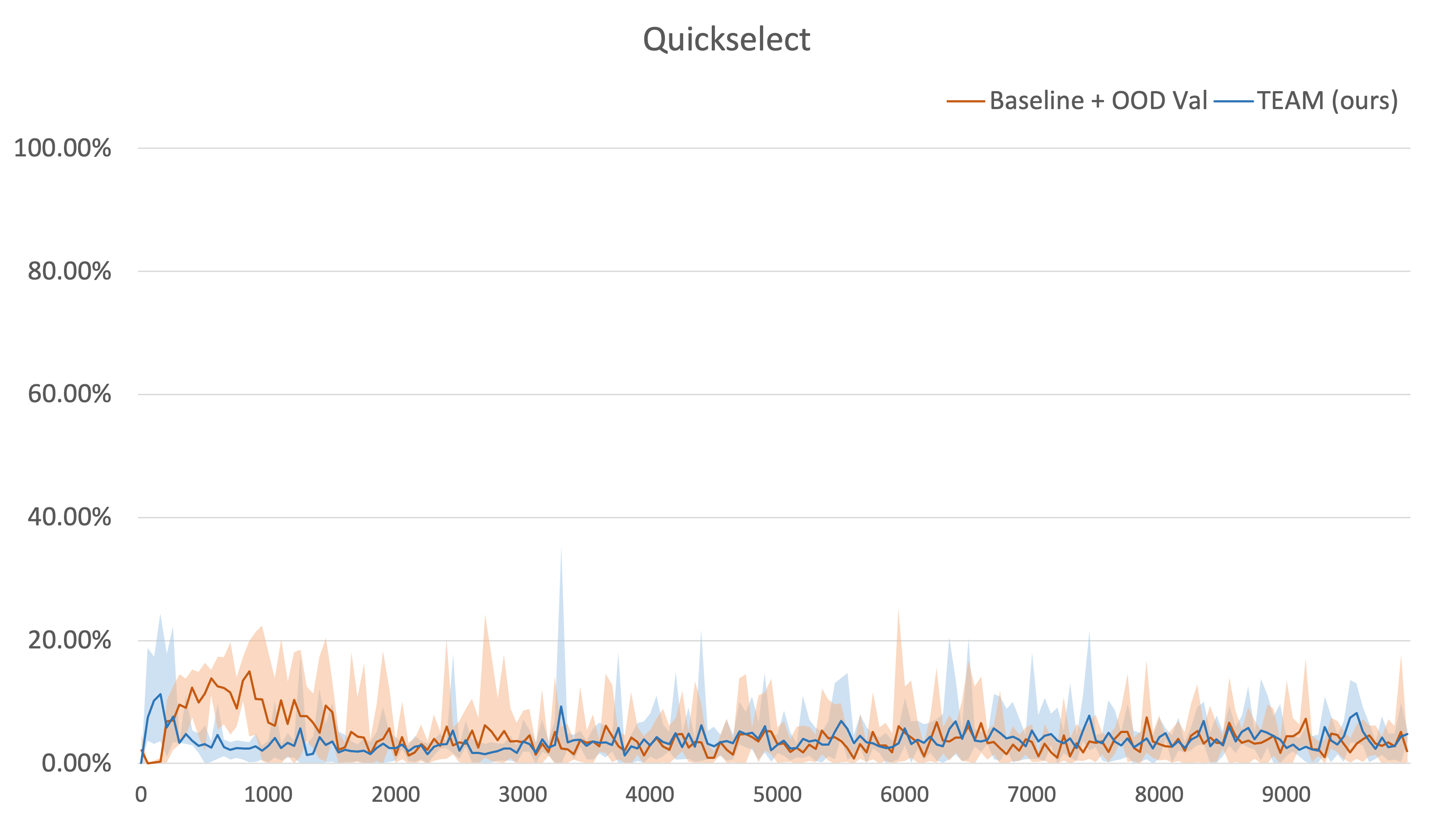}
    \end{subfigure}
    \begin{subfigure}[b]{0.245\linewidth}
        \centering
        \includegraphics[width=\textwidth]{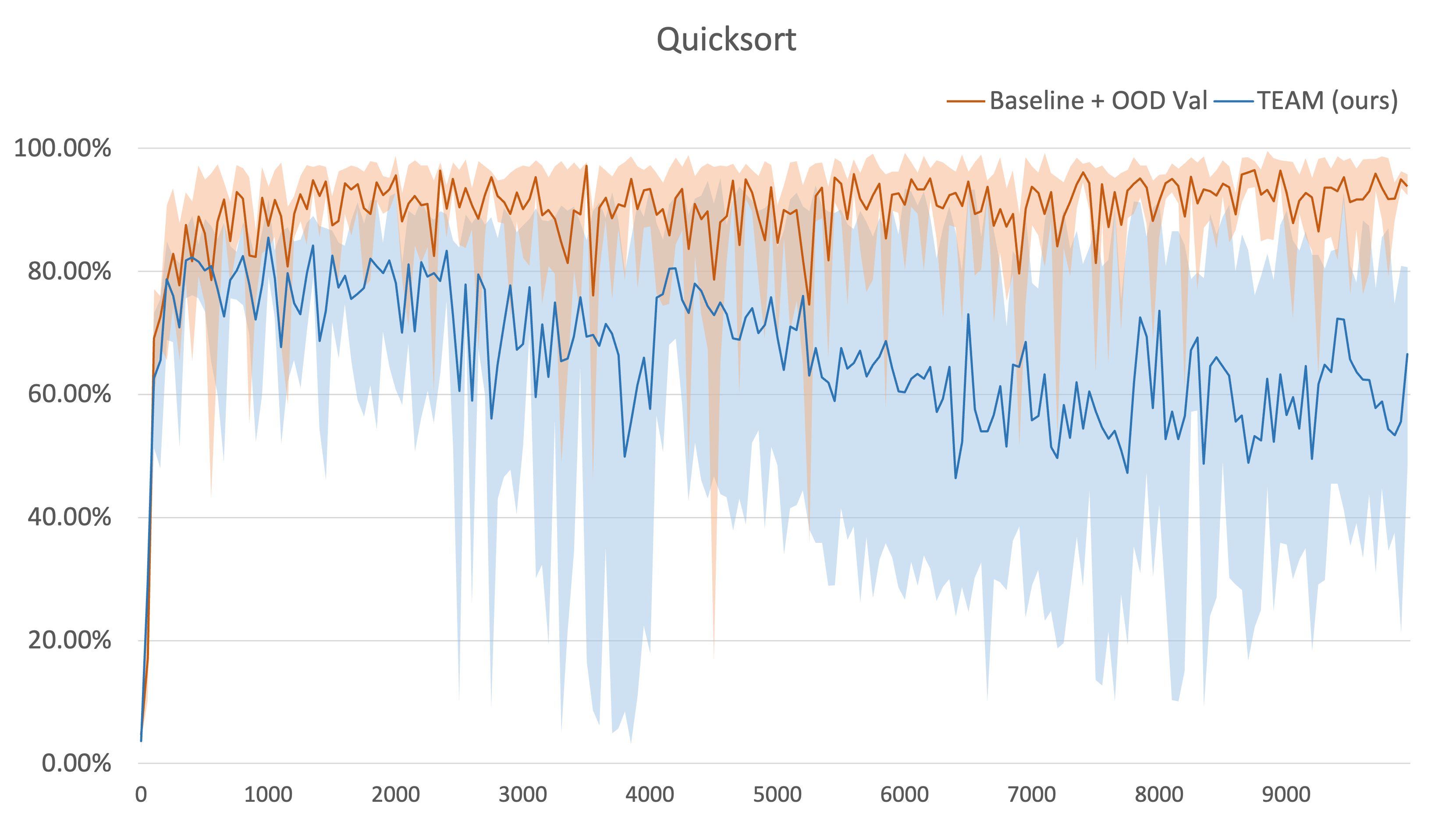}
    \end{subfigure}
    \begin{subfigure}[b]{0.245\linewidth}
        \centering
        \includegraphics[width=\textwidth]{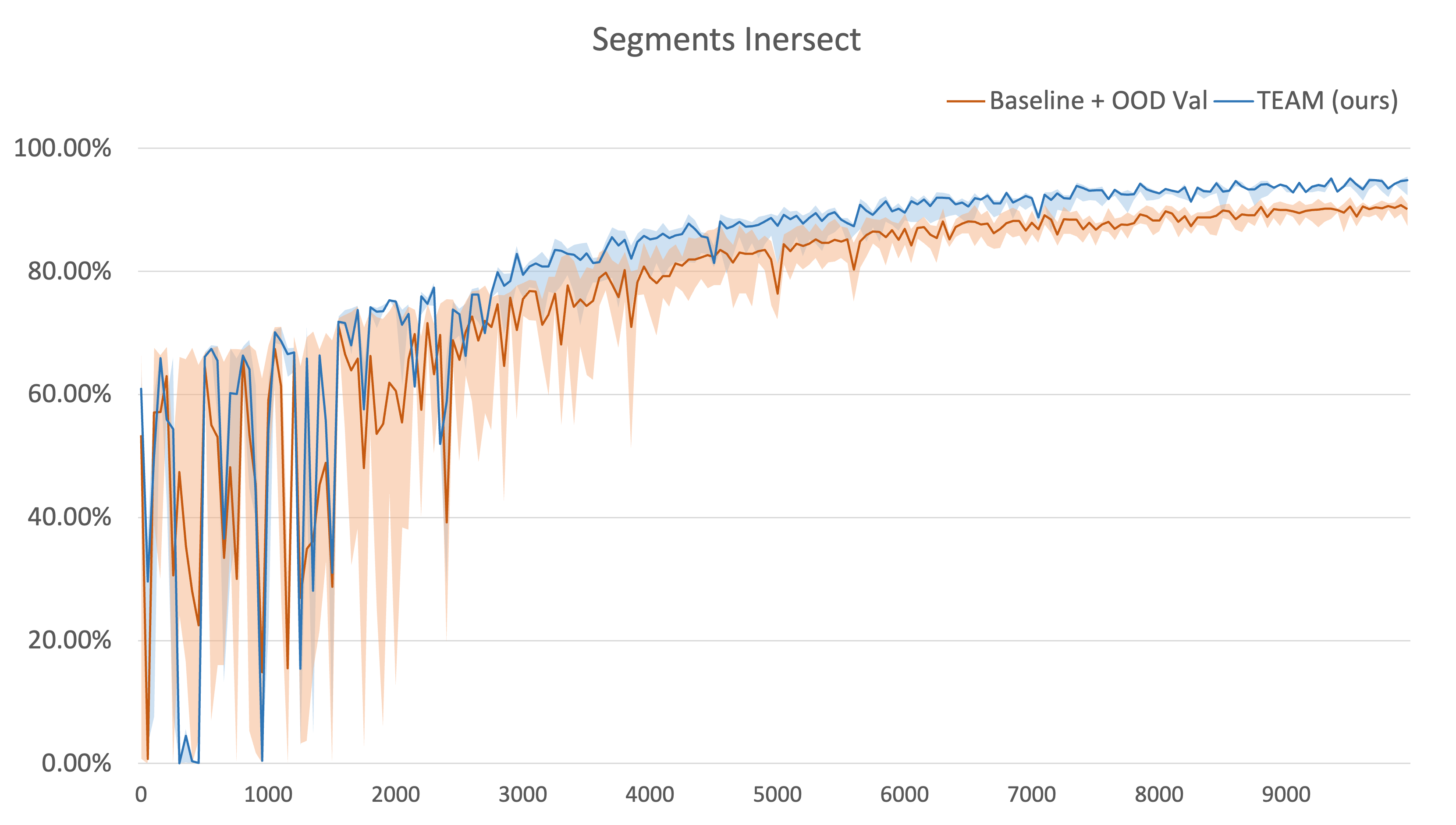}
    \end{subfigure}
    \begin{subfigure}[b]{0.245\linewidth}
        \centering
        \includegraphics[width=\textwidth]{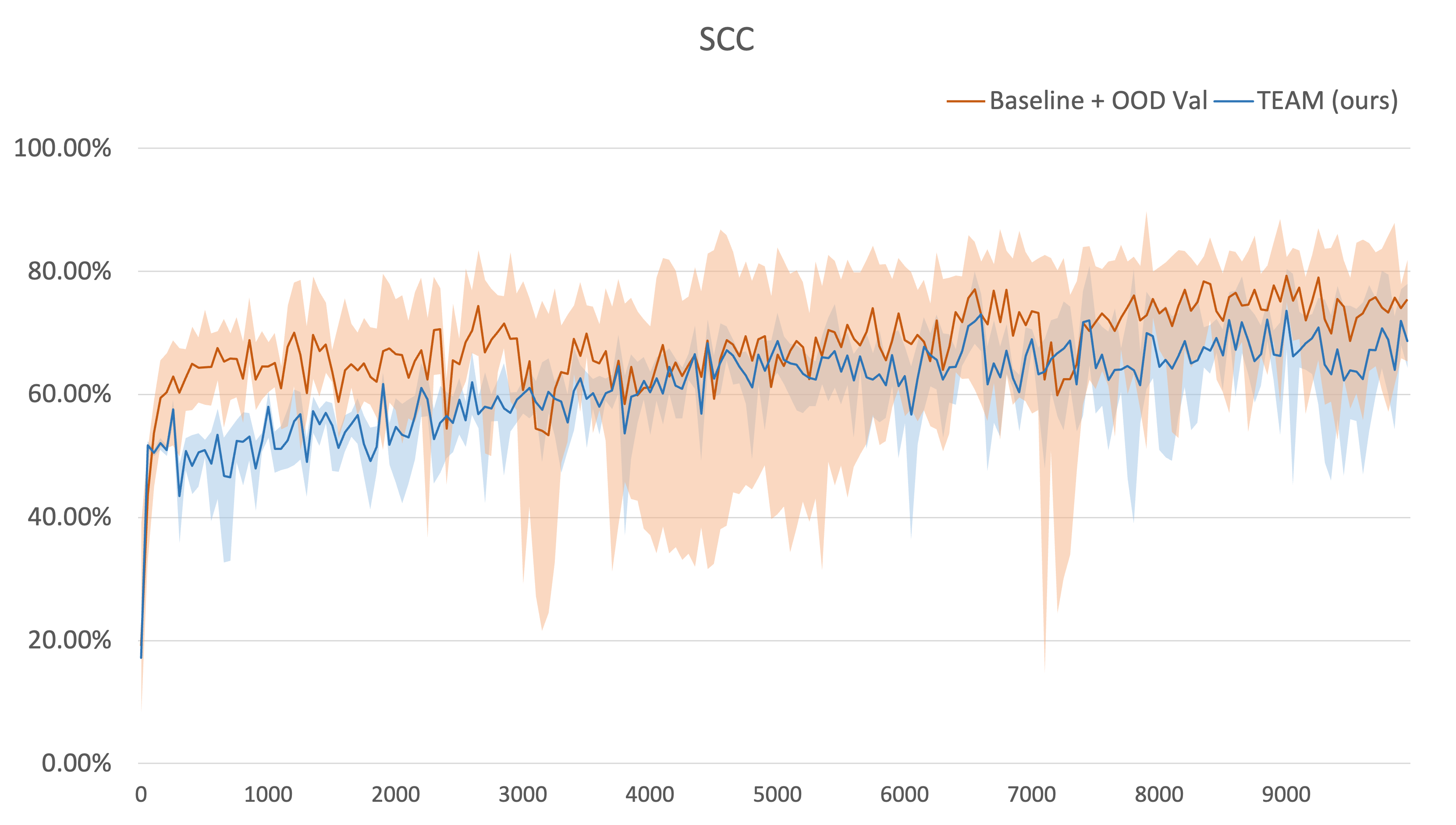}
    \end{subfigure}
    \begin{subfigure}[b]{0.245\linewidth}
        \centering
        \includegraphics[width=\textwidth]{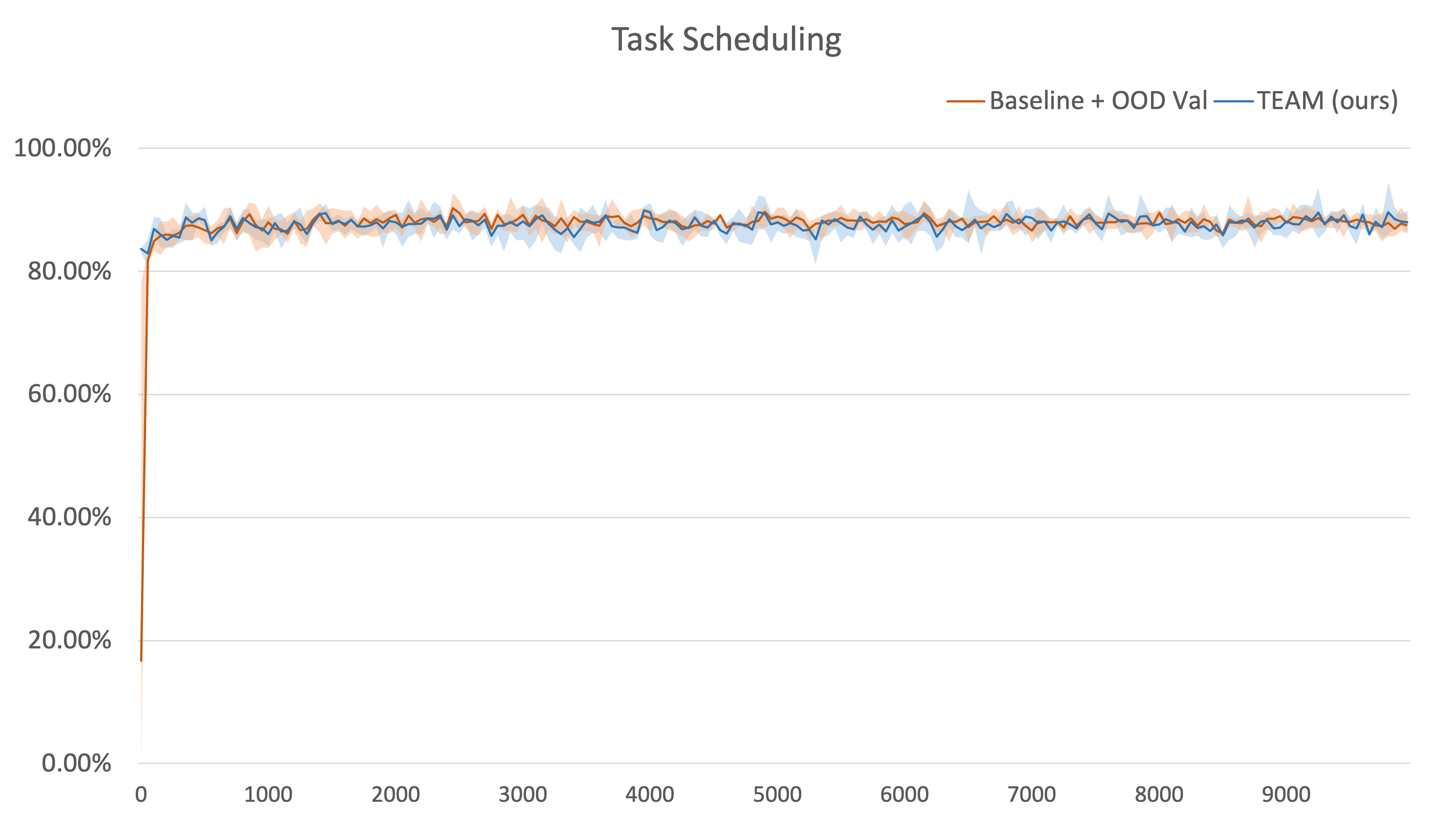}
    \end{subfigure}
    \begin{subfigure}[b]{0.245\linewidth}
        \centering
        \includegraphics[width=\textwidth]{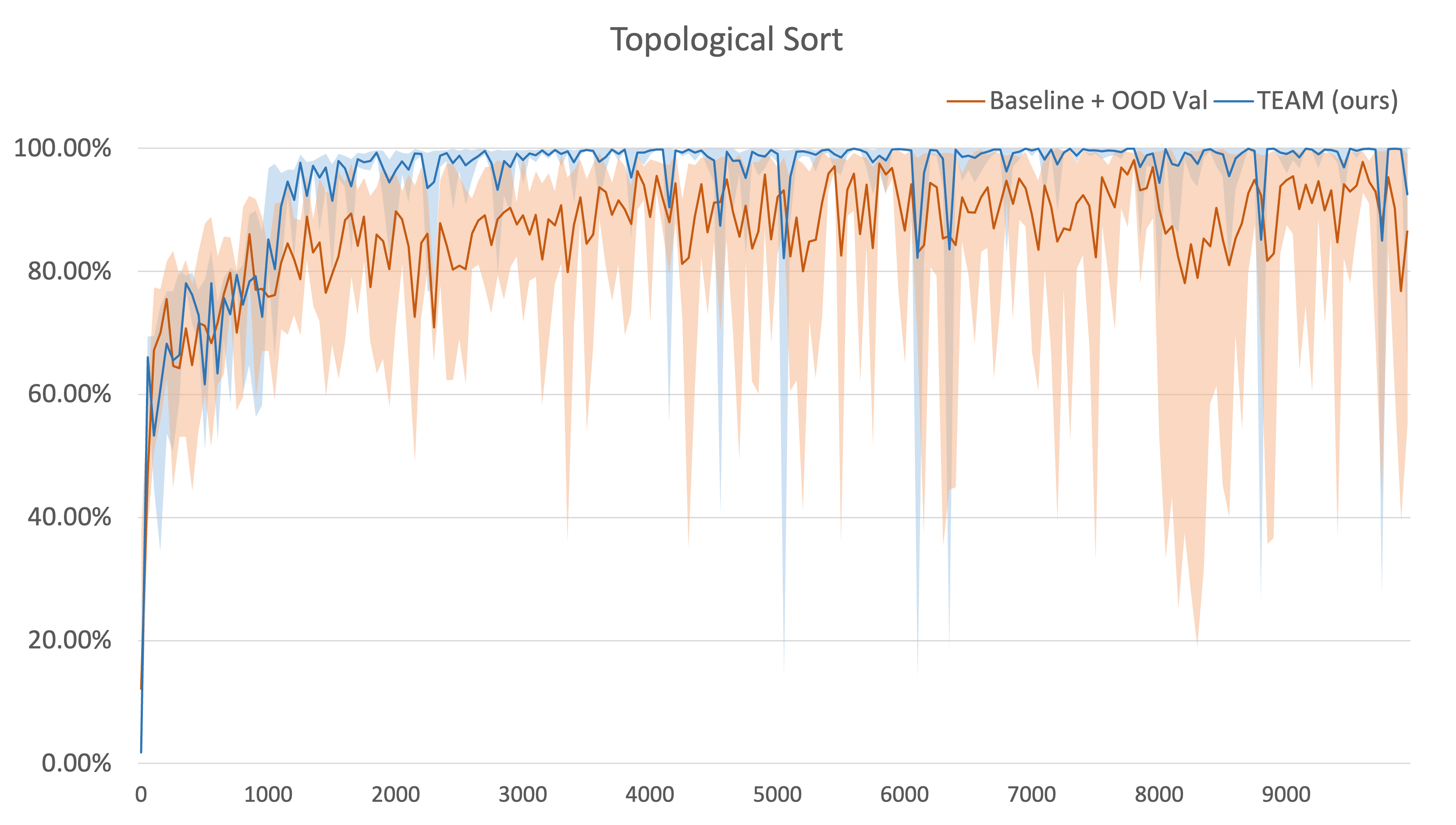}
    \end{subfigure}
        \caption{The learning curve of each algorithm, where the blue line indicates the learning curve of our TEAM model, and the orange line indicates the learning curve of the baseline model with OOD validation. The algorithms are ordered in alphabetical order, the same as the Table \ref{tab:30}}
        \label{fig:learning_curve}
\end{figure}

\begin{table}[h]
	\centering
	\caption{Average OOD micro-F1 scores of the baseline (Triplet-GMPNN), baseline with OOD validation set, and our best TEAM model, after 10,000 training steps on single algorithm tasks.}
	\label{tab:30}
	\begin{tabular}{lccc}
		\toprule
		Algorithm & Baseline~\cite{ibarz:2022genaralist}  & Baseline + OOD Val & TEAM (Ours)\\
            \midrule
    		Activity Selector      & ${95.18\%\pm0.45}$     & ${94.60\%\pm1.19}$     & ${95.90\%\pm1.61}$\\
    		Articulation Points    & ${88.32\%\pm2.01}$     & ${54.35\%\pm9.56}$     & ${86.89\%\pm7.03}$\\
    		Bellman-Ford           & ${97.39\%\pm0.19}$     & ${95.35\%\pm1.20}$     & ${96.08\%\pm0.48}$\\
            BFS                    & ${99.73\%\pm0.04}$     & ${99.58\%\pm0.19}$     & ${99.85\%\pm0.09}$\\
            Binary Search          & ${77.58\%\pm2.35}$     & ${86.27\%\pm5.49}$     & ${85.31\%\pm4.42}$\\
            Bridges                & ${93.99\%\pm2.07}$     & ${91.87\%\pm5.68}$     & ${97.53\%\pm1.39}$\\
            Bubble Sort            & ${67.68\%\pm5.50}$     & ${84.26\%\pm10.30}$    & ${74.35\%\pm14.15}$\\
            DAG Shortest Paths     & ${98.19\%\pm0.30}$     & ${93.75\%\pm1.33}$     & ${98.13\%\pm0.27}$\\
            DFS                    & ${47.79\%\pm4.19}$     & ${61.97\%\pm16.66}$    & ${59.31\%\pm27.75}$\\
            Dijkstra               & ${96.05\%\pm0.60}$     & ${93.87\%\pm1.08}$     & ${93.14\%\pm0.93}$\\
            Find Max. Subarray     & ${76.36\%\pm0.43}$     & ${65.70\%\pm3.72}$     & ${69.79\%\pm1.60}$\\
            Floyd-Warshall         & ${48.52\%\pm1.04}$     & ${42.56\%\pm2.15}$     & ${35.33\%\pm2.82}$\\
            Graham Scan            & ${93.62\%\pm0.91}$     & ${90.77\%\pm4.25}$     & ${94.12\%\pm1.61}$\\
            Heapsort               & ${31.04\%\pm5.82}$     & ${47.95\%\pm23.13}$    & ${54.42\%\pm8.85}$\\
            Insertion Sort         & ${78.14\%\pm4.64}$     & ${91.11\%\pm2.22}$     & ${90.73\%\pm2.89}$\\
            Jarvis’ March          & ${91.01\%\pm1.30}$     & ${85.22\%\pm10.84}$    & ${93.02\%\pm1.91}$\\
            Knuth-Morris-Pratt     & ${19.51\%\pm4.57}$     & ${62.85\%\pm24.11}$    & ${62.59\%\pm23.28}$\\
            LCS Length             & ${80.51\%\pm1.84}$     & ${86.38\%\pm0.26}$     & ${85.67\%\pm0.27}$\\
            Matrix Chain Order     & ${91.68\%\pm0.59}$     & ${95.48\%\pm0.75}$     & ${94.46\%\pm0.88}$\\
            Minimum                & ${97.78\%\pm0.55}$     & ${97.67\%\pm1.13}$     & ${99.23\%\pm0.13}$\\
            MST-Kruskal            & ${89.80\%\pm0.77}$     & ${86.96\%\pm3.25}$     & ${78.47\%\pm2.36}$\\
            MST-Prim               & ${86.39\%\pm1.33}$     & ${89.24\%\pm1.21}$     & ${93.30\%\pm1.74}$\\
            Naive String Matcher   & ${78.67\%\pm4.99}$     & ${87.81\%\pm14.98}$    & ${99.90\%\pm0.15}$\\
            Optimal BST            & ${73.77\%\pm1.48}$     & ${84.64\%\pm0.43}$     & ${70.70\%\pm5.45}$\\
            Quickselect            & ${0.47\%\pm0.25}$      & ${1.73\%\pm0.83}$      & ${3.84\%\pm3.56}$\\
            Quicksort              & ${64.64\%\pm5.12}$     & ${64.98\%\pm11.66}$    & ${55.49\%\pm15.60}$\\
            Segments Intersect     & ${97.64\%\pm0.09}$     & ${92.15\%\pm0.13}$     & ${94.96\%\pm0.21}$\\
            SCC                    & ${43.43\%\pm3.15}$     & ${47.14\%\pm5.89}$     & ${44.30\%\pm3.43}$\\
            Task Scheduling        & ${87.25\%\pm0.35}$     & ${85.56\%\pm1.18}$     & ${87.70\%\pm2.17}$\\
            Topological Sort       & ${87.27\%\pm2.67}$     & ${78.08\%\pm24.27}$    & ${100.00\%\pm0.01}$\\
            \midrule
            Overall Avg.           & ${75.98\%}$            & ${78.00\%}$            & ${79.82\%}$\\
		\bottomrule
	\end{tabular}
\end{table}

\begin{table}[h]
	\centering
	\caption{The ratio of the number of parameters between the baseline model (Triplet-GMPNN) and our TEAM model.}
	\label{tab:params}
	\begin{tabular}{lc}
		\toprule
            Algorithm & TEAM (Ours) / Baseline~\cite{ibarz:2022genaralist} \\
            \midrule
            Activity Selector      & ${1.46}$\\
            Articulation Points    & ${1.17}$\\
            Bellman-Ford           & ${1.17}$\\
            BFS                    & ${1.17}$\\
            Binary Search          & ${1.46}$\\
            Bridges                & ${1.16}$\\
            Bubble Sort            & ${1.17}$\\
            DAG Shortest Paths     & ${1.07}$\\
            DFS                    & ${1.11}$\\
            Dijkstra               & ${1.17}$\\
            Find Max. Subarray     & ${1.46}$\\
            Floyd-Warshall         & ${1.14}$\\
            Graham Scan            & ${1.26}$\\
            Heapsort               & ${1.11}$\\
            Insertion Sort         & ${1.17}$\\
            Jarvis’ March          & ${1.46}$\\
            Knuth-Morris-Pratt     & ${1.26}$\\
            LCS Length             & ${1.44}$\\
            Matrix Chain Order     & ${1.14}$\\
            Minimum                & ${1.46}$\\
            MST-Kruskal            & ${1.26}$\\
            MST-Prim               & ${1.17}$\\
            Naive String Matcher   & ${1.46}$\\
            Optimal BST            & ${1.14}$\\
            Quickselect            & ${1.26}$\\
            Quicksort              & ${1.17}$\\
            Segments Intersect     & ${1.46}$\\
            SCC                    & ${1.11}$\\
            Task Scheduling        & ${1.46}$\\
            Topological Sort       & ${1.11}$\\
            \midrule
            Overall Avg.           & ${1.25}$\\
		\bottomrule
	\end{tabular}
\end{table}


\end{document}